# Resilient Supplier Selection in Logistics 4.0 with Heterogeneous Information

**Md Mahmudul Hasan**
Department of Mechanical and Industrial Engineering,
Northeastern University
360 Huntington Ave, Boston, MA 02115, USA
email: hasan.mdm@husky.neu.edu

**Dizuo Jiang**
Department of Mechanical and Industrial Engineering,
Northeastern University
360 Huntington Ave, Boston, MA 02115, USA
email: jiang.di@husky.neu.edu

**A. M. M. Sharif Ullah**
School of Regional Innovation and Social Design Engineering,
Kitami Institute of Technology
165 Koen-cho, Kitami, Hokkaido 090-8507, Japan
email: ullah@mail.kitami-it.ac.jp

**Corresponding author**
**Md. Noor-E-Alam***
Department of Mechanical and Industrial Engineering,
Northeastern University
360 Huntington Ave, Boston, MA 02115, USA
*Corresponding author email: mnalam@neu.edu

**Abstract:** Supplier selection problem has gained extensive attention in the prior studies. However, research based on Fuzzy Multi-Attribute Decision Making (F-MADM) approach in ranking resilient suppliers in logistic 4.0 is still in its infancy. Traditional MADM approach fails to address the resilient supplier selection problem in logistic 4.0 primarily because of the large amount of data concerning some attributes that are quantitative, yet difficult to process while making decisions. Besides, some qualitative attributes prevalent in logistic 4.0 entail imprecise perceptual or judgmental decision relevant information, and are substantially different than those considered in traditional suppler selection problems. This study develops a Decision Support System (DSS) that will help the decision maker to incorporate and process such imprecise heterogeneous data in a unified framework to rank a set of resilient suppliers in the logistic 4.0 environment. The proposed framework induces a triangular fuzzy number from large-scale temporal data using probability-possibility consistency principle. Large number of non-temporal data presented graphically are computed by extracting granular information that are imprecise in nature. Fuzzy linguistic variables are used to map the qualitative attributes. Finally, fuzzy based TOPSIS method is adopted to generate the ranking score of alternative suppliers. These ranking scores are used as input in a Multi-Choice Goal Programming (MCGP) model to determine optimal order allocation for respective suppliers. Finally, a sensitivity analysis assesses how the Supplier's Cost versus Resilience Index (SCRI) changes when differential priorities are set for respective cost and resilience attributes.

**Keywords:** Logistic 4.0, Resilience, Supplier selection, Supplier's Cost versus Resilience Index (SCRI), TOPSIS, Fuzzy Multi-Attribute Decision Making (F-MADM)

## 1. Introduction

With the increasing requirements for industrial process due to technological revolution, companies are facing rigorous challenges such as competitive global industry, increasing market unpredictability, swelling customized product demands and shortened product renewal cycle (Hofmann & Rüsch, 2017). Aligned with the demanding industry, the fourth wave of technological innovation has been materialized, known as Industry 4.0. It refers to so-called fourth industrial revolution in discrete and process manufacturing, logistics and supply chain (Logistics 4.0), energy (Energy 4.0) etc., which is resulted due to the digital transformation of industrial markets (industrial transformation) integrated with smart manufacturing. Powered by foundational technology such as autonomous robots, simulation, cyber security, the cloud and additive manufacturing (Rüßmann et al., 2015), Industry 4.0 improves the manufacturing systems to an intelligent level that takes advantage of advanced information and manufacturing technologies to achieve flexible, smart, and reconfigurable manufacturing processes in order to address a dynamic and global market (Zhong, Xu, Klotz, & Newman, 2017).

As an essential and significant part of Industry 4.0, Logistics 4.0 concerns the various aspects of end-to-end logistics in the context of Industry 4.0, the Internet of Things (IoT), cyber-physical systems, automation, big data, cloud computing, and Information technology (Hofmann & Rüsch, 2017; i-SCOOP, 2017a; Rüßmann et al., 2015; Zhong et al., 2017). Logistics 4.0 aims to develop a smart logistics system to fulfill the customer requirements in the current connected, digitalized and rapidly changing global logistics market (i-SCOOP, 2017b). To adopt with an Industry 4.0 environment, extensive cutting-edge applications have been landed within logistics 4.0. Juhász and Bányai (2018) identified challenges of just-in-sequence supply in the automotive industry from the aspect of Industry 4.0 solutions and detected impacts of Industry 4.0 paradigm on just-in-sequence supply. Ivanov, Dolgui, Sokolov, Werner, and Ivanova (2016) proposed a dynamic model and algorithm for short-term supply chain scheduling problem that simultaneously considered both machine structure selection and job assignments in smart factories. Brettel, Friederichsen, Keller, and Rosenberg (2014)visualized the supply chain process by introducing cyber-physical systems to bridge the advanced communication between machines on the landscape of Industry 4.0.

With higher priority on customer satisfaction, incremental attention has been drawn to the availability, reliability, flexibility and agility of the logistics system (Barreto, Amaral, & Pereira, 2017; Witkowski, 2017). Unpredictable natural catastrophes or unexpected man-made disasters, such as earthquakes, floods, labor strikes and bankruptcy engender serious threat to the capability of the logistics system on these aspects. Despite of low occurrence probability, the tremendous financial impacts of the disruptions in any form on the logistics system are much more obvious. Renesas Electronics Corporation, a Japanese semiconductor manufacturer and the world's largest manufacturer of microcontrollers is an industry 4.0 company. They launched the R-IN32M4-CL2 industrial Ethernet communication specific standard product (ASSP) with integrated Gigabit PHY to support the increasing network and productivity for Industry 4.0 companies on June 25, 2015 (Corporation, 2015). In earthquake and tsunami that struck the northeast coast of Japan on March 2011, the corporation's Naka Factory and other manufacturing facilities were severely damaged by the earthquake. The total losses caused by the disaster was 814.2 million USD, even though the insurance covered 198.9 million USD (Ye & Abe, 2012). UPS, the leading logistics company, which has achieved great advancement in the fields of digitalized logistics system and smart factory in the wave of Industry 4.0, suffers from the lack of resilient supply chain as well. During the hurricane Florence that threatened the east coast in mid-September 2018, the delivery rate reduced to 50% with thousands of delivery exceptions (SUPPLYCHAINDIVE, 2018).

To withstand against disruption a resilient supplier is indispensable in the sourcing decision process while operating under the principles of logistics 4.0. A resilient supplier usually has high adaptive capability to reduce the vulnerability against disruptions, absorb disaster impact and quickly recover from disruption to ensure desired level of continuity in operations following a disaster (Y. Sheffi & Rice Jr, 2005; Y. J. M. P.

B. Sheffi, 2005). To comprehensively evaluate the alternatives and select the optimal supplier, diverse factors are needed to be taken into consideration. Several proactive strategies such as suppliers' business continuity plans, fortification of suppliers, maintaining contract with back up suppliers, single and multiple sourcing, spot purchasing, collaboration and visibility are considered to enhance supply chain resilience in the presence of operational and disruption risks (Namdar, Li, Sawhney, & Pradhan, 2018; Torabi, Baghersad, & Mansouri, 2015). Another study suggests that suppliers' reliability, and flexibility in production capacity play key role in developing contingency plans to help mitigate the severity of disruptions (Kamalahmadi & Mellat-Parast, 2016). Criteria such as supply chain complexity, supplier resource flexibility, buffer capacity and responsiveness were considered in traditional resilient supplier selection problem (Haldar, Ray, Banerjee, & Ghosh, 2012, 2014). However, to select the resilient supplier in logistics 4.0 environment, additional aspects are required to be taken into consideration regarding the features of logistics 4.0.

In logistics 4.0, most of the companies and related organizations are adopting end-to-end information sharing technologies because the amount of data produced and shared through the supply chain is significantly increasing (Domingo Galindo, 2016; G. Wang, Gunasekaran, Ngai, & Papadopoulos, 2016). Driven by the extensive practical and efficient attributes, numerous applications of big data are employed in delivery forecasting, optimal routing, and productivity monitoring and labor reduction (Waller & Fawcett, 2013). Making well-informed decisions in sourcing process also involves a variety of massive data-based logistics aspects evaluation such as delivery lead time, inventory level, production capacity, and operational investment. Information sources such as ERP transaction data, GPS-enabled data, machine-generated data and RFID data are frequently transferred, stored and retrieved through the logistics system, which require massive data manipulation methodology (Rozados & Tjahjono, 2014).

With the widespread use of IoT, cyber-physical systems, and Information technology, attention has been given on the supplier's performance and ability in responding to changing customer demand with agility, warehouse automation, logistics system digitalization, information management and IT security, etc. The relevant data from these fields are generally collected in various formats at quick velocity, and entails large volume—all-together leads to Big Data, which is often available as real-time and historical data. Data that are collected in real-time is characterized by time series whereas historical data is often presented in graphical format. Attributes that entail large amount of information in logistics 4.0 are inventory level, schedule of delivery, production capacity, cost etc. Processing these large amounts of data requires a formal computation process that can enable decision maker to efficiently evaluate alternative decisions.

Therefore, for selecting resilient suppliers in logistics 4.0, a decision-making framework is needed given the extensive impact of sourcing decision on the supply chain resiliency, efficiency and sustainability. The decision problem involves evaluation of several alternative suppliers against multiple conflicting criteria.

Moreover, this problem even becomes more complicated when the decision relevant information (DRI) are in heterogeneous form such as qualitative information that are imprecise in nature and vague sometimes, and large number of quantitative information that are difficult to process. To address this problem, we propose a Decision Support System (DSS) leveraging the principle of Multi-Attribute Decision Making (MADM) to rank alternative suppliers from resilience and logistic 4.0 perspectives.

The key contributions of this study are:

i. As the existing research on MADM has limited applicability in the logistics 4.0 environment, we, for the first time extend the F-MADM framework to logistics 4.0 industries where selecting resilient suppliers has far reaching consequence. The proposed DSS is capable of handling qualitative attributes that entails imprecise DRI, and are substantially different than those considered in traditional supplier selection problem. Moreover, our plan is to integrate large number of quantitative DRI, which in logistics 4.0 environment is characterized by time series and graphical information. As such, we propose an integrated decision-making framework to process this heterogeneous information in a seamlessly unified framework to facilitate the resilient supplier selection process for a logistics 4.0 industry.

ii. Commonly used fuzzy based TOPSIS technique solely depends on the standard triangular linguistic class to handle the qualitative appraisal in the decision making process and lacks the sophistication in processing the quantitative information. Our proposed DSS overcomes this limitation by successfully converting and integrating the crisp granular information extracted from graphically presented data into triangular fuzzy based TOPSIS decision matrix. Crisp granular or c-granular information refers to the pieces of information entailing well–defined crisp or ill-defined perceptual boundary consisting of sharp numbers (Ullah & Noor-E-Alam, 2018; Lotfi A. Zadeh, 1997).

iii. Selecting suppliers by considering all attributes that share equal preference usually generate inflated set of ranking score which is generic, however, sometimes fails to address the issue if a decision maker wants to put more importance on one set of criteria than another. We divide the attributes in two sets—mainly based on measure of resilience and cost—used as efficiency measure. We further demonstrate the capability of the proposed DSS to generate Supplier's Cost versus Resilience Index (SCRI) based on customized preference given by the decision makers on cost and resilience attributes.

iv. We also extend the proposed DSS with the help of an order allocation model leveraging Multi-Choice Goal Programming (MCGP) technique. The optimal suppliers identified with the help of F-MADM approach are only suitable for single-sourcing problems in which the procurement quantity can be satisfied by a single supplier. However, in a situation where procurement demand cannot be fulfilled by a single supplier, and strategic decision makers want to diversify their market, stabilize their sourcing channels while distributing risks on multiple suppliers and drive up competitiveness, allocating orders among competitive suppliers can turn out to be a viable strategy. As such, our framework will empower decision makers to allocate orders among alternative suppliers by taking into account the ranking of individual suppliers that has been generated via F-MADM approach.

The rest of this paper is organized as follows. In Section 2, we review the relevant literatures. Then in section 3 and 4, we present the theoretical concept needed to design the decision-making framework. Section 5 describes the proposed DSS for suppler evaluation and order allocation problem. In Section 6, we illustrate the effectiveness of the proposed DSS via a case study. Finally, in Section 7 we conclude the work conducted in this study with future research directions.

## 2. Literature review

Multi-Criteria Decision Analysis (MCDA) approaches are widely adopted in the fields of transportation, immigration, education, investment, environment, energy, defense and healthcare (Devlin, Sussex, & Economics, 2011; Dodgson, Spackman, Pearman, & Phillips, 2009; Gregory et al., 2012; Mühlbacher, Kaczynski, & policy, 2016; Nutt, King, & Phillips, 2010; Wahlster, Goetghebeur, Kriza, Niederländer, & Kolominsky-Rabas, 2015). Howard and Ralph (Raiffa & Keeney, 1975) first introduce MCDA as a methodology for evaluating alternatives based on individual preference, often against conflicting criteria, and combining them into one single appraisal. Prior studies have also applied MCDA approach to select suppliers using multiple attributes (Lo & Liou, 2018; Ren, Xu, & Wang, 2018; Sodenkamp, Tavana, & Di Caprio, 2018). Multi-Criteria Decision Making (MCDM) with grey numbers was used to propose a conceptual framework for suppliers’ management entailing selection, segmentation and development of resilient suppliers (Valipour Parkouhi, Safaei Ghadikolaei, & Fallah Lajimi, 2019). They used Grey DEMATEL technique to weigh the criteria considered for the two dimensions of resilience enhancer and resilience reducer. Finally, Grey Simple Additive Weighting (GSAW) technique was used to determine the ranking score of each supplier according to each dimension.

As one of the most prevalent MCDA approaches, Analytical Hierarchy Process (AHP) has been widely adopted to address supplier selection problems (De Felice, Deldoost, & Faizollahi, 2015; Prasad, Prasad, Rao, & Patro, 2016). Saaty (1980) at first proposed the methodology of AHP, which was then refined by Golden, Wasil, and Harker (1989). In AHP method, the feature of original data set is usually qualitative. In decision-making process, the master problem is decomposed to sub-problems, making the unidirectional hierarchical relationships between levels more understandable. Based on the subdivisions, pairwise comparison between alternatives is conducted to determine the importance of the criteria and priority over all alternatives. During this decision-making process, the evaluation of alternatives is extended to qualitative field while multiple criteria are considered, and the consistency of the system is satisfied.

However, due to the subjectivity of the qualitative information resulted from the discrepancy of decision makers' experience, knowledge and judgment, the uncertainty and imprecise nature in the data are not dealt with, which may impair the reliability and robustness of the result.

To help the stakeholders establish a more accurate and reliable approach, Yoon (Yoon, 1987) and Hwang et al. (Hwang, Lai, & Liu, 1993) developed TOPSIS (Technique for order preference by similarity to an ideal solution). The underlying idea is that the optimal solution should have the closest distance from the Positive Ideal Solution (PIS) and longest distance from the Negative Ideal Solution (NIS). TOPSIS can handle quantitative input data, which is different from the basic feature of AHP. Because of its precise nature, TOPSIS has been broadly applied in supplier selection problem. Shahroudi and Tonekaboni (2012) adopted TOPSIS in the supplier selection process in Iran Auto Supply Chain, in which both the numerical and linguistic evaluation criteria are considered to determine the preferential alternatives. In this study, numerical numbers (without consideration of fuzziness of data set) are assigned to qualitative data directly to generate the quantitative decision matrix for TOPSIS. To develop an integrated decision-making framework, some group of researchers aggregated AHP with TOPSIS to better evaluate the alternative suppliers (Bhutia & Phipon, 2012; Şahin & Yiğider, 2014). However, most of these methods utilized crisp information, and thus uncertainty, impreciseness and fuzziness nature of the judgmental information are not considered.

To obtain better results in problems where decision making and analysis are significantly affected by the uncertainty inherent in the DRI, the fuzzy technique was introduced. Gan, Zhong, Liu, and Yang (2019) used fuzzy Best-Worst Method (BWM) to determine the decision makers' weight and modular TOPSIS to sort and rank alternative suppliers form resiliency perspective in a random and group decision making framework. Haldar et al. (2014) integrated Triangular and trapezoidal linguistic data to select resilient suppliers using TOPSIS. However, the criteria used in these two studies were based on traditional supply chain, which are not sufficient to comprehensively evaluate suppliers from the perspective of logistics 4.0 and resiliency, simultaneously. Moreover, while evaluating suppliers they did not consider the quantitative decision relevant information, which is often available for several attributes considered in case of logistics 4.0.

K. T. Atanassov (1999), at first defined the concept and properties of Intuitionistic fuzzy set, which was then adopted by Boran, Genç, Kurt, and Akay (2009) and to the aggregated decision-making framework in supplier selection problem. H. Wang, Smarandache, Sunderraman, and Zhang (2005) and Haibin, Smarandache, Zhang, and Sunderraman (2010) proposed the concept of *single valued neutrosophic set* (SVNS), which can characterize the indeterminacy of a perceptual information more explicitly. SVNS was then aggregated with TOPSIS by Şahin and Yiğider (2014) to replace the crisp information in the decision matrix. Their findings show that TOPSIS when integrated with SVNS performs better with incomplete, undetermined and inconsistent information in MCDA problems. As most of the membership functions in the research mentioned above are assumed to be triangular, to find another way to capture the vagueness of the qualitative information, Positive Trapezoidal Fuzzy Number (PTFN) was proposed by Bohlender, Kaufmann, and Gupta (1986) and was introduced by Herrera and Herrera-Viedma (2000) in group decision making problems. C.-T. Chen, Lin, and Huang (2006) adopted PTFN to present a fuzzy decision-making framework to deal with supplier selection problem.

However, in case of SVNS, when the decision makers' evaluation are provided as a single number within the interval [0,1], it does not necessarily represent the underlying uncertainty associated with that evaluation scheme. Thus, in such context it is preferred to represent the decision makers' assessment by an interval rather than a single number, indicating to the significance of using Inter-valued Fuzzy Sets (IVFS). Guijun and Xiaoping (1998) defined the concept of IVFS, while Ashtiani, Haghighirad, Makui, and ali Montazer (2009) extend the application of IVFS in TOPSIS to solve Multi Criteria Decision Making problems. Foroozesh, Tavakkoli-Moghaddam, and Mousavi (2017) developed a multi-criteria group decision making model integrating IVFS and fuzzy possibilistic statistical concepts to weigh the decision makers involved in decision making process. Finally, with the help of a relative-closeness coefficient based technique, they rank resilient suppliers under the interval-valued fuzzy uncertainty. Additionally, K. Atanassov and Gargov (1989) proposed the notion of Interval-valued Intuitionistic Fuzzy Sets (IVIFS) as a further generalization of fuzzy set theory. Lakshmana Gomathi Nayagam, Muralikrishnan, and Sivaraman (2011) adopted IVIFS in multi criteria decision making problem. The method was then extended by T.-Y. Chen, Wang, and Lu (2011) and T.-Y. Chen (2015) to group decision making setting , while Mohammad (2012) implemented this approach in supplier selection problem. After reviewing the relevant literatures, we found that the state-of-the-art studies deal with only qualitative attributes, while we argue that many of the essential and significant evaluation attributes may entail quantitative DRI, especially for selecting resilient suppliers for logistics 4.0 companies.

It is true that the existing research have shown promising potential of MCDA methods and fuzzy techniques in supplier selection problems (Gan et al., 2019; Haldar et al., 2012, 2014; Hasan, Shohag, Azeem, Paul, & Management, 2015; Jiang, Faiz, & Hassan, 2018). However, there are limitations of the state-of-the-art literatures: (i) to the best of our knowledge, no existing research extends F-MADM framework in supplier evaluation problems leveraging large number of information (time series and graphical information), (ii) no prior study investigated F-MADM approach for evaluating suppliers' performance from resilience and logistics 4.0 perspective, simultaneously, and (iii) it is not clear how fuzzy based TOPSIS can be extended to process inherent uncertainty in decision relevant information associated with both the quantitative and qualitative attributes. These gaps in the existing studies create an avenue for further research to extend F-MADM framework to help decision makers in logistics 4.0 industries to strategically select resilient suppliers considering qualitative and large number of quantitative information, which lies in the central focus of this study.

## 3. Multi Attribute Decision Making (MADM) and fuzzy logic

### 3.1 MADM

MADM provides a comprehensive decision analysis framework that could help the stakeholders balance the advantages and disadvantages of the alternatives in a multi-dimensional optimization problem, in which alternatives and evaluation attributes are the essential variables. The general decision-analysis procedure of MADM and the corresponding steps are summarized in Table 1 according to (Thokala et al., 2016):

Table 1.
Framework of MADM

| | | |
|---|---|---|
| Step 1 | Defining the decision problem | Select optimal supplier with highest resilience over a group of alternative suppliers |
| Step 2 | Selecting and structuring attributes | Identify the evaluation attributes with respect to supplier resilience |
| Step 3 | Measuring performance | Gather data about the alternatives' performance on the attributes and summarize this in a decision matrix |
| Step 4 | Scoring alternatives | Evaluate the performance of the alternative suppliers based on the objective of the attributes |
| Step 5 | Weighting criteria and decision makers | Determine the weight of attributes and decision makers based on their importance |
| Step 6 | Calculating aggregate scores | Use the alternatives' scores on the attributes and the weights for the attributes and decision makers to get "total value" by which the alternatives are ranked with TOPSIS |
| Step 7 | Dealing with uncertainty | Perform Sensitivity analysis to understand the level of robustness of the MADM results |
| Step 8 | Reporting and examination of findings | Interpret the MADM outputs, including sensitivity analysis, to support decision making |

### 3.2 Technique for order preference by similarity to an ideal solution (TOPSIS)

TOPSIS is a decision-making technique wherein the alternatives are evaluated based on their numerical distance to the ideal solution. The closer the distance of an alternative to the ideal solution and the farther to the negative ideal solution, the higher a grade it would obtain. Because in this study, we are adopting triangular possibility distribution or Triangular Fuzzy Number (TFN) to express the performance of the alternatives, Euclidian Distance is used to measure the performance of the alternatives, and the function is described as below (Şahin & Yiğider, 2014; Singh, 2016):

$$s_i^+ = \sqrt{\sum_{j=1}^{n} \left\{ \left(a_{ij} - a_j^+\right)^2 + \left(b_{ij} - b_j^+\right)^2 + \left(c_{ij} - c_j^+\right)^2 \right\}} \quad i = 1,\ 2,\ \ldots,\ n \tag{3.1}$$

$$s_i^- = \sqrt{\sum_{j=1}^{n} \left\{ \left(a_{ij} - a_j^-\right)^2 + \left(b_{ij} - b_j^-\right)^2 + \left(c_{ij} - c_j^-\right)^2 \right\}} \quad i = 1,\ 2,\ \ldots,\ n \tag{3.2}$$

$$\tilde{\rho}_i = \frac{s_i^-}{s_i^+ + s_i^-}, \quad 0 \le \tilde{\rho}_i \le 1 \tag{3.3}$$

Where $s_i^+$ and $s_i^-$ are the positive and negative ideal solution respectively, $\tilde{\rho}_i$ is the closeness coefficient , $a_{ij}$, $b_{ij}$, $c_{ij}$ are the component of the TFN that express the performance of alternatives on criteria *j*. $a_j^+$, $b_j^+$, $c_j^+$ are the corresponding components of the Positive Ideal Solution (PIS) and $a_j^-$, $b_j^-$, $c_j^-$ are the corresponding components of Negative Ideal Solution (NIS).

**3.3 Properties of Triangular Fuzzy Number (TFN)**

A Triangular Fuzzy Number (TFN) shown in Figure 1 is defined with three points as follows:

$$\tilde{A} = (a, b, c)$$

Where [a, c] is the support and $\mu_{\tilde{A}}(b) = 1$ is the core of the fuzzy number. This representation is interpreted in terms of membership functions as follows:

$$\mu_{\tilde{A}}(x) = \begin{cases} 0, & for\ x < a \\ \frac{x-a}{b-a}, & for\ a < x < b \\ \frac{c-x}{c-b}, & for\ b < x < c \\ 0, & for\ c < x \end{cases} \tag{3.4}$$

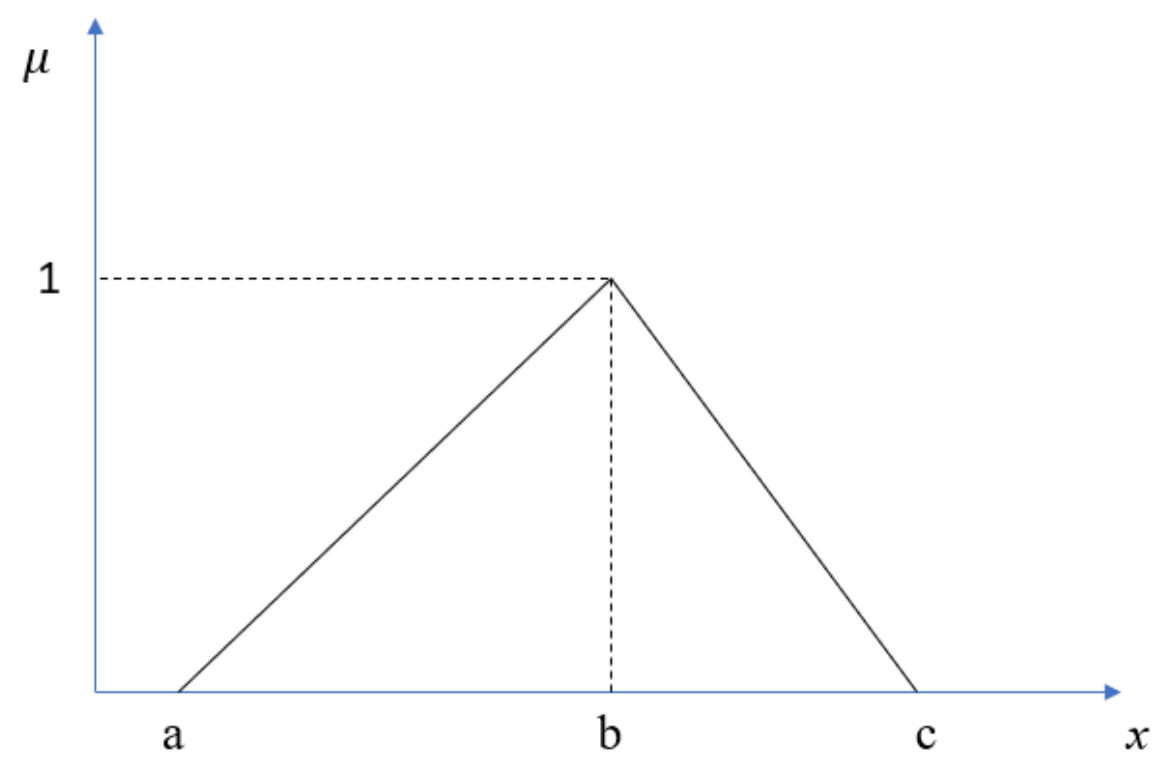

Figure 1. Triangular Fuzzy Number

The operation of the TFN could be summarized as follows (Mahapatra, Mahapatra, & Roy, 2016): let $A = (a_1, b_1, c_1)$, $B = (a_2, b_2, c_2)$, r is a real number, then,

1) Addition: $A + B = (a_1 + a_2, b_1 + b_2, c_1 + c_2)$
2) Subtraction: $A - B = (a_1 - c_2, b_1 - b_2, c_1 - a_2)$ (3.5)
3) Multiplication: $A \times B = (a_1 \times a_2, b_1 \times b_2, c_1 \times c_2)$

$$A \times r = (a_1 \times r, b_1 \times r, c_1 \times r)$$

The underlying reason for using TFN instead of other fuzzy techniques is primarily due to the probability-possibility consistency principle that induces TFN from time series data. Moreover, TFN also provides the basis for converting granular information that are extracted from graphically presented historical data.

### 3.4 Membership function and reliability modification

The definition of membership function was first introduced by L. A. Zadeh (1965), where the membership functions were used to operate on the domain of all possible values. In fuzzy logic, membership degree represents the truth value of a certain proposition.

Different from the concept of probability, truth value represents membership in vaguely defined sets. For any set X, the membership degree of an element x of X in fuzzy set A is denoted as $\mu_A(x)$, which quantifies the grade of membership of the element x to the fuzzy set A. To calculate the membership degree, the universe of discourse concerning different attributes are fuzzified using linguistic classes according to the granule definiteness axiom of multi-granularity (A.M.M. Sharif Ullah, 2005; Ullah, 2005). Figure 2 shows the fuzzification of universe of discourse or frame of discernment consisting [a, b].

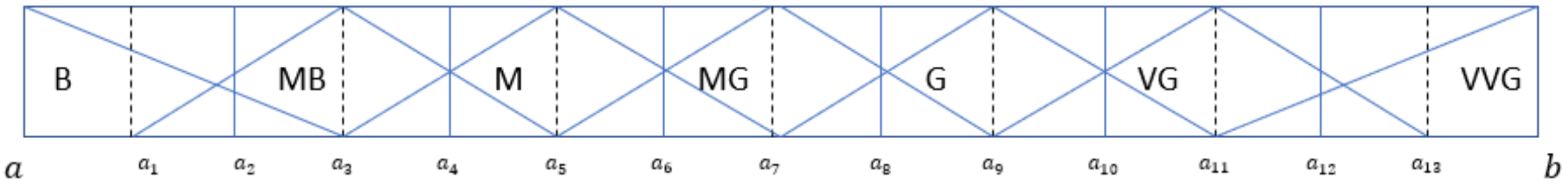

Figure 2. Fuzzified frame of discernment

For the frame of discernment shown in the Figure 2, the membership functions for the 7 different classes (B, MB, …,VVG) are calculated as follows:

$$m_B = max\left(0,\ \frac{a_3 - x}{a_3 - a}\right)$$

$$m_{MB} = \max\left(0,\ \min(\frac{x-a_1}{a_3-a_1}, \frac{a_5-x}{a_5-a_3}\right) \quad (3.6)$$

……

$$m_{VVG} = \max\left(0,\ \frac{x - a_{11}}{b - a_{11}}\right)$$

where $a_i = a + \frac{b-a}{2\times 7}$, $i = 1,2,\dots,(2\times 7 - 1)$, and 7 is the number of class in this frame of discernment. The membership functions are assumed to be triangular and symmetric. The membership function for each class depends on the frame of discernment of the attribute.

As the membership functions are assumed triangular and symmetric for the fuzzified frame of discernment, the uncertainty and impreciseness of the functions need to be taken into consideration. Wen, Miaoyan, and Chunhe (2017) proposed a reliability-based modification to deal with uncertainty of information and the reliability of information sources. The reliability of the membership functions is measured by the static reliability index and dynamic reliability index. Static reliability index is defined by the similarity among classes, while dynamic reliability index is measured by the risk distance between the test samples and the overlapping area among classes, respectively. The comprehensive reliability is computed by the product of the two index, and the reliability-based membership function are fused using Dempster's combination rule (Dempster, 1967; Wen et al., 2017). The numerical examples provided by Jiang et al. (2018) verified the effectiveness of the reliability modification approach in membership functions.

The static reliability index is measured by the overlapped area between two adjacent classes. In Figure 3, the shaded area is the overlapped region between classes M and MG.

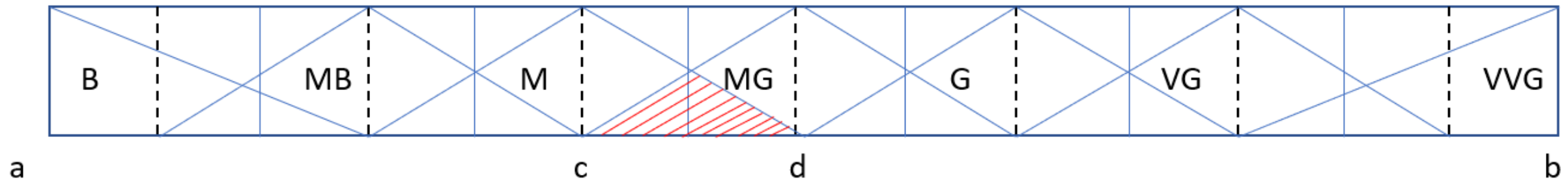

Figure 3. Illustration of static reliability index

The larger the overlapped area between classes M and MG, the more likely that an input data is wrongly recognized in a linguistic class. The similarity between classes M and MG $sim_{M,MG}$ in a certain attribute and the corresponding static reliability index $R_j^s$ for $C_j$ can be described according to Wen et al. (2017):

$$sim_{M,MG} = \frac{\int_c^d \min\limits_{c\le x\le d}(m_M(x), m_{MG}(x))dx}{\int m_M(x) + m_{MG}(x) - \int_c^d \min\limits_{c\le x\le d}(m_M(x), m_{MG}(x))dx} \tag{3.7}$$

$$R_j^s = \sum_{i<l}(1 - sim_{il}) \tag{3.8}$$

where $i$ and $l$ are the adjacent classes in the same universe of discourse in one attribute.

The dynamic reliability index is measured with a set of test sample and calculated by the risk distance between the peak of overlap area and the test value.

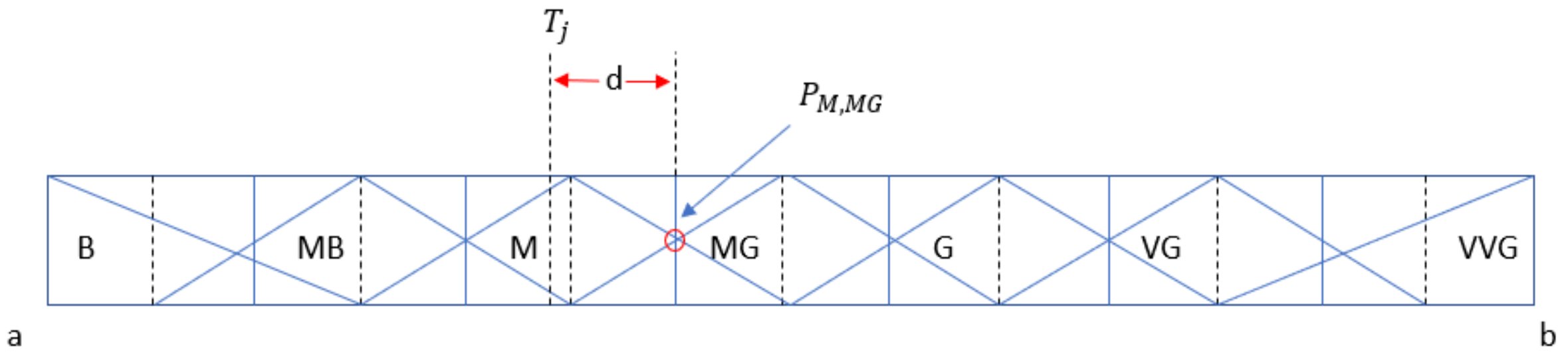

Figure 4. Illustration of dynamic reliability index

If $P_{M,MG}$ is the peak of the overlap area between classes M and MG in Figure 4, and $T_j$ is the test sample generated for $C_j$, the distance $d$ between $T_j$ and $P_{M,MG}$ represents the risk distance that related to the uncertainty of the test sample. The risk distance and dynamic reliability index for $C_j$ can be formulated as:

$$d_{M,MG} = \frac{|\mathrm{T}_j - P_{M,MG}|}{D} \tag{3.9}$$

$$R_j^d = e^{\sum_2^n d_{(l-1)l}} \tag{3.10}$$

where D is the range of the universe of discourse of $\mathrm{C}_j$, which is $(\mathrm{a} - \mathrm{b})$ in Figure 3.

Then the comprehensive reliability index for $C_j$ can be defined as:

$$R_j = R_j^s \times R_j^d \tag{3.11}$$

After the normalization we get,

$$R_j^* = \frac{R_j}{\max(R_j)} \tag{3.12}$$

Then, the reliability-modified membership degree can be calculated as:

$$m_{jl}^{R_j^*} = R_j^* \times m_l \tag{3.13}$$

where *l* is a linguistic class in a universe of discourse.

## 4. Quantitative data analytics

### 4.1 Processing time-series data

A. M. M. Sharif Ullah and Shamsuzzaman (2013) proposed an approach that can represent the uncertainty under a large set of continuous time-series input parameters (temporal data) by point cloud and transfer it to a graphical fuzzy number based on probability-possibility transformation. The transformation process is generalized as follows:

Assume, we have a temporal data presented in time-series data as shown in Figure 5.

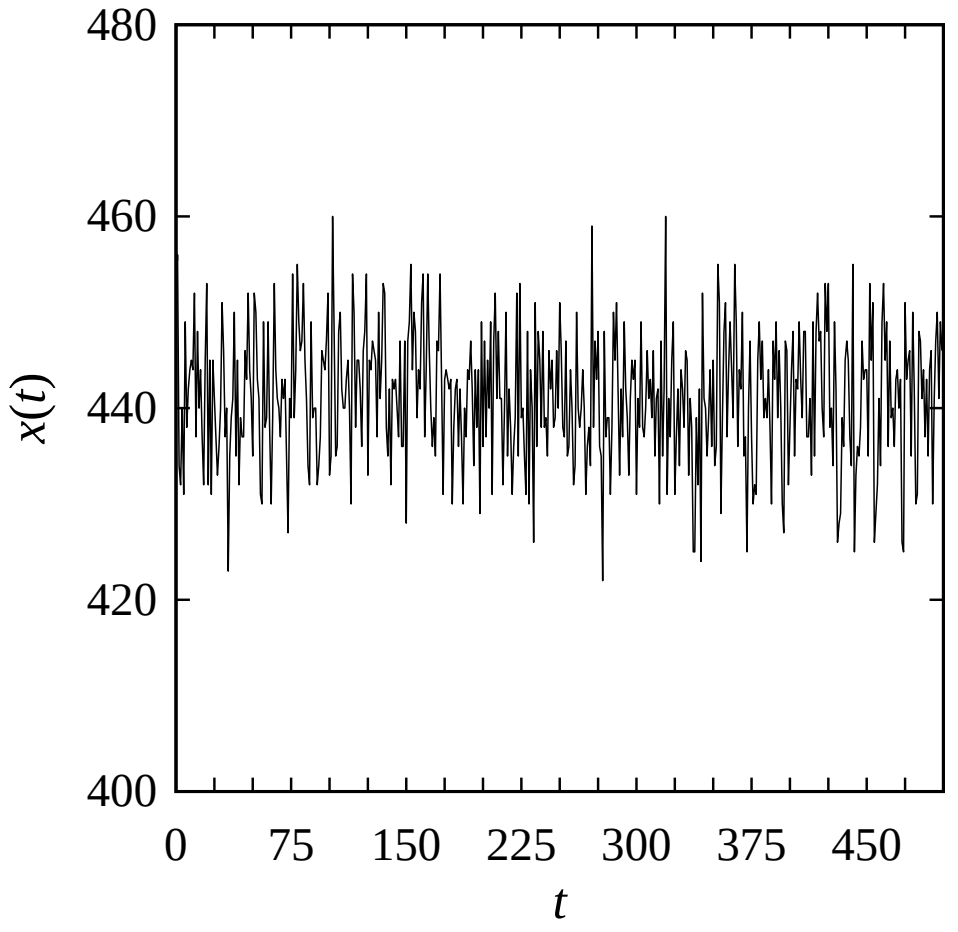

Figure 5. Original time-series data

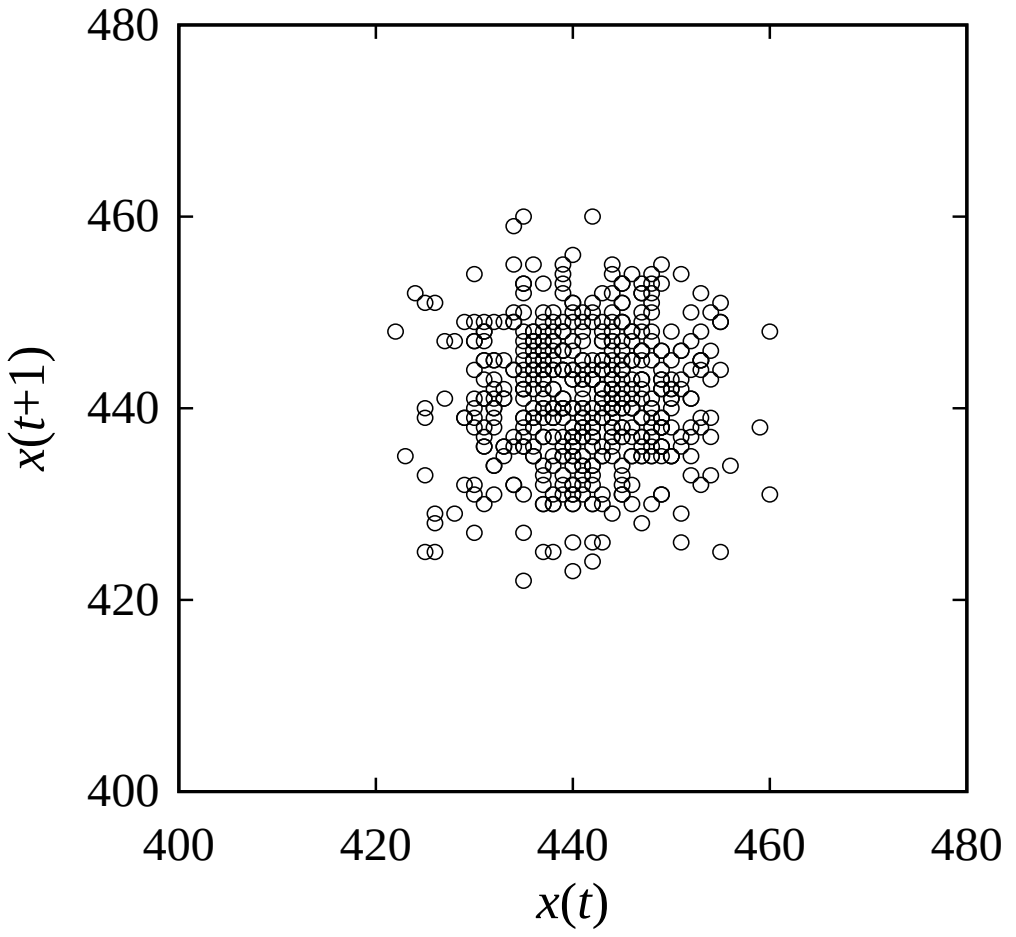

Figure 6. Transferred point-cloud

If we set the $x(t)$ as the x-coordinate and $x(t+1)$ as the y-coordinate, this set of parameters could be represented as a point cloud as shown in Figure 6, providing a visual/computational representation of variability, modality, and ranges associated with the quantity.

Assume, $g(x)$ be the probability density functions (pdf) that represent the underlying point-cloud of $x(t)$, the cumulative pdf, $F(x)$, can be defined as:

$$F(x) = \int g(x)dx$$

Let $PrA(x)$ denote the following formulation:

$$PrA(x) = \frac{dF(x)}{dx}$$

A possibility distribution given by the membership function $\mu(x)$ can be defined as:

$$\mu(x) = \frac{PrA(x)}{\max(PrA(x) \mid \forall\, x \in X)}$$

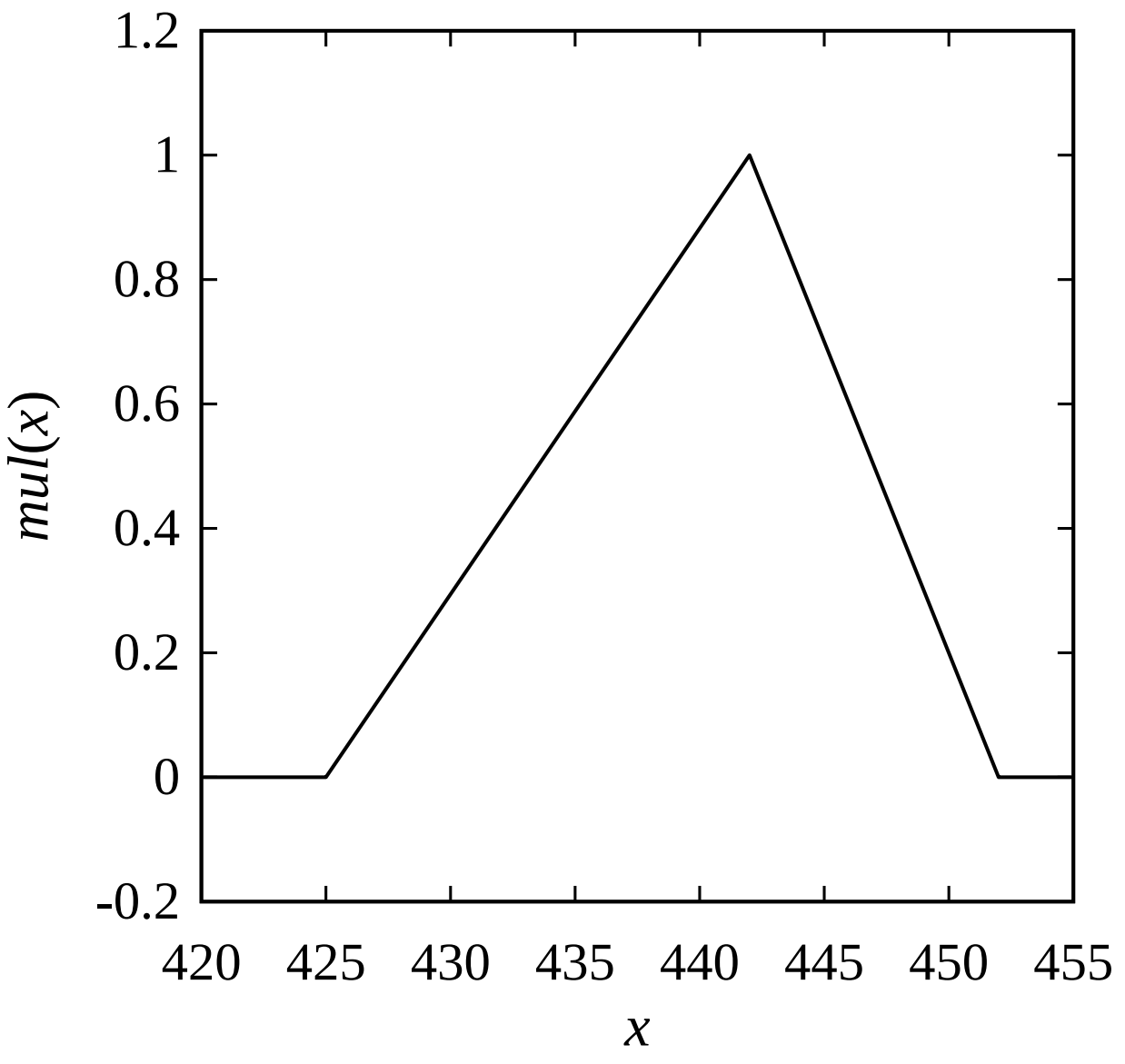

Figure 7. Graphical Triangular Fuzzy Number.

After the probability-possibility transformation, the point cloud is transferred to a triangular possibility distribution or TFN in a graphical format as shown in Figure 7. In what follows, the triangular fuzzy set can be expressed as:

$$A = (425, 442, 452)$$

## 4.2 Processing graphical information

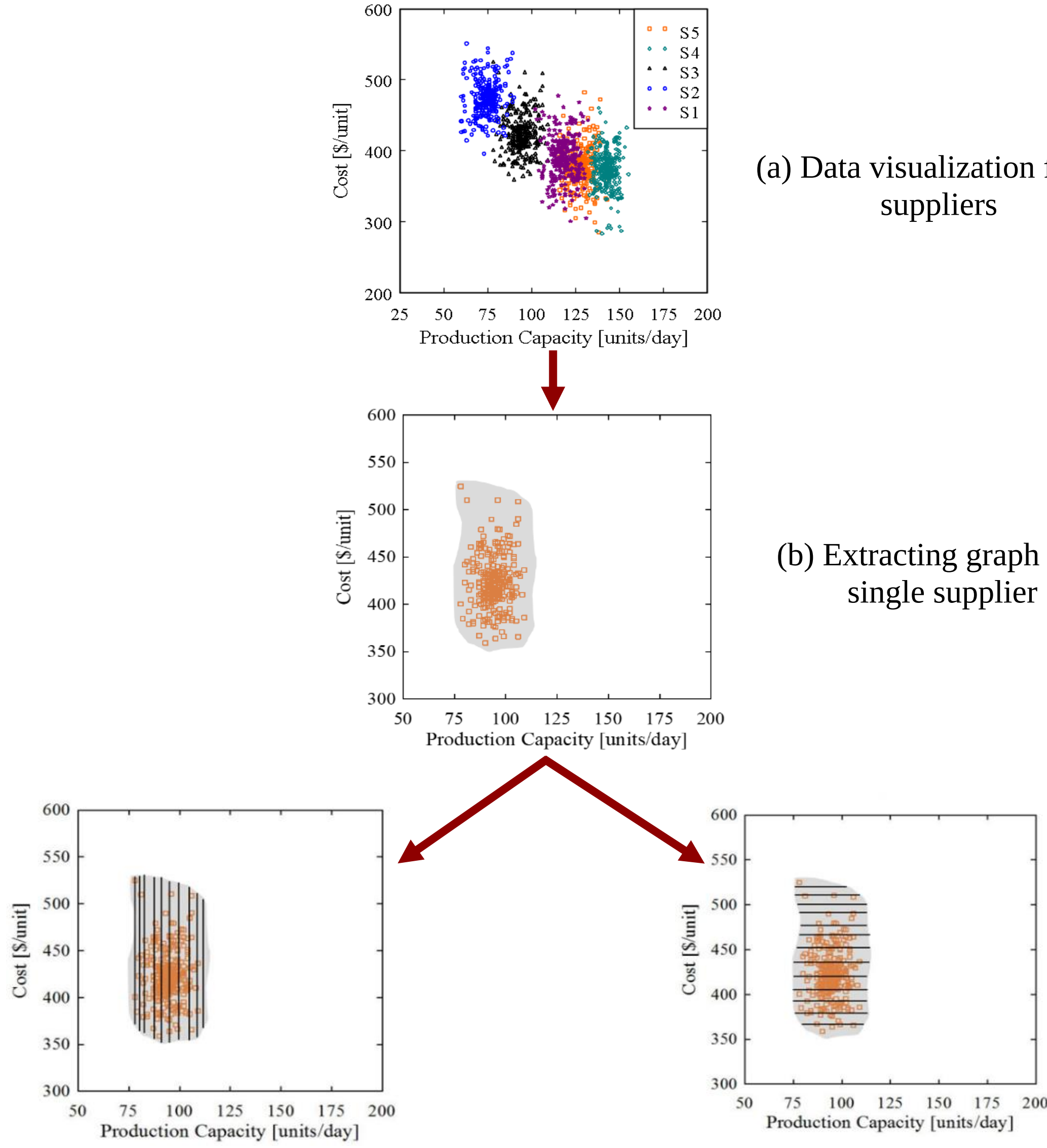

(a) Data visualization for all suppliers

(b) Extracting graph for a single supplier

(c) Crisp granular information

(d) Crisp granular information concerning production capacity

Figure 8. Graphical information extraction process.

In the logistics 4.0 system, the data are not always generated in the format of continuous time-series. Sometimes they are generated discretely and stored in the data base for future access. These historically stored data can be expressed in pieces of graphs (Domingo Galindo, 2016) because when visualized, these pieces of graphs can potentially present a large amount of information in an easy-to-understand way. Then, the necessary DRI in terms of pieces of crisp granular information can be extracted from these graphically presented data according to (Ullah & Noor-E-Alam, 2018). Figure 8(a) shows the visualization of data collected for five suppliers concerning cost per unit of product and production capacity. Then, graphical data associated with a single supplier can be extracted from the combined graph and is presented in Figure 8(b). Once data for individual supplier is presented graphically, then the associated DRI concerning cost per unit of product and production capacity can be extracted in the form of crisp granular information or ranges as visualized in Figure 8(c) and Figure 8(d), respectively.

## 5.0 Proposed supplier evaluation and order allocation model

In this Supplier Evaluation and Order Allocation Model, both the quantitative and qualitative information are characterized by Triangular Fuzzy Number (TFN) to evaluate the performance of the alternatives in a unified platform, simultaneously. For the quantitative attributes, time-series and non-time series data are transferred to fuzzy set through point cloud and graphic extraction approaches, respectively, while qualitative assessments, including performance and weight appraisal, are transferred directly based on the standard fuzzification process of the frame of discernment. With the fuzzy set for all attributes and weights, the weighted decision matrix is constructed. Then the algorithm of TOPSIS is performed to evaluate the performance of the alternatives and generate the list of preference based on the obtained ranking score. Finally, the ranking score are regarded as the coefficient in the MCGP to calculate the order allocation plan that best fulfill the requirements of the decision makers. To provide a comprehensive and understandable illustration for the proposed supplier evaluation and order allocation model, we present a complete computation process with detailed description below:

***Step 1: Processing Quantitative data***

In this step, we process and transfer the large number of quantitative data entailing time-series and non-time series data, which are available in pieces of graphics. Following two sub-steps constitute this step.

***Step 1(a): Transferring time-series based quantitative data to TFN***

At first, the time-series data, for supplier $S_i$ on attribute $C_j$ are expressed by a point cloud in the form of $P(x(t), x(t+1))$ according to the principle explained in section 4.1. After the point cloud transformation, the data are transferred to a possibility distribution of triangular form as shown in Figure 7, which can be represented by a TFN in the form of $A_{ij}\ (a_{ij}, b_{ij}, c_{ij})$.

***Step 1(b): Transferring non-time series based graphical data to TFN***

After the information extracted from the graphical information in the form of crisp granular information, the randomly obtained $r$ number of crisp granular information or ranges for supplier $S_i$ on attribute $C_j$, $R_{ijr}(p_{ijr}, q_{ijr})$, are presented in Table 2.

Table 2.
Example of extracted ranges

| Ranges | $C_j$ |
|---|---|
| $R_{ij1}$ | $(p_{ij1}, q_{ij1})$ |
| $R_{ij2}$ | $(p_{ij2}, q_{ij2})$ |
| $\cdots$ | $\cdots$ |
| $R_{ijr}$ | $(p_{ijr}, q_{ijr})$ |

After collecting all the crisp granular information for every alternative supplier, we fuzzify the frame of discernment associated with every non-time series attribute $C_j$ based on the fuzzification approach proposed in (Ullah & Noor-E-Alam, 2018). In the fuzzification process, the span of the frame of discernment is generated by the minimum and maximum of all the extracted crisp granular information for a certain attribute regarding all the alternative suppliers, while the number of linguistic terms is determined according to the granule definiteness axiom (A.M.M. Sharif Ullah, 2005; Ullah, 2005). For Example, if $p_{min} = \min_r p_{ijr}$, $q_{max} = \max_r q_{ijr}$, the frame of discernment of $C_j$ presented as $U = [p_{min}, q_{max}]$ can be fuzzified as shown in Figure 9.

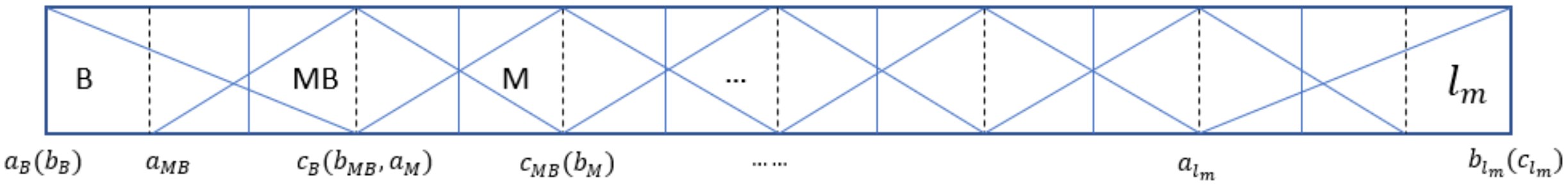

Figure 9. Fuzzified frame of discernment.

After the fuzzification process, the linguistic classes $l$ and associated TFN table are constructed as in Table 3:

Table 3.
Linguistic terms and associated TFN

| Linguistic Terms | $TFN\ (a, b, c)$ | | |
|---|---|---|---|
| $B$ | $a_B$ | $b_B$ | $c_B$ |
| $MB$ | $a_{MB}$ | $b_{MB}$ | $c_{MB}$ |
| $\cdots$ | $\cdots$ | $\cdots$ | $\cdots$ |
| $l_m$ | $a_{l_m}$ | $b_{l_m}$ | $c_{l_m}$ |

Where B and MB represent linguistic terms expressed as Bad, Moderately Bad, and $l_m$is $m^{th}$ linguistic term that can assume the form such as Bad (B), Moderately Bad (MB), Moderately Good (MG), Good (G) etc.

With the fuzzified frame of discernment, the membership degree for every range value $R_{ijr}(p_{ijr}, q_{ijr})$ on each linguistic class is computed based on (3.6).

$$m_{ij} = \frac{\int_{x\in R} m_F(x)dx}{\|\acute{R}\|} \tag{5.1}$$

where R refers to the span of the criteria and $\|\acute{R}\|$ refers to the largest segment of R that belongs to the support $m_F$. This way, the membership degree of $R_{ijr}(p_{ijr}, q_{ijr})$ on attribute $C_j$ at linguistic class $l$ is calculated as $M_{ijrl}$.

Because all the membership functions are assumed to be symmetric and triangular, we perform reliability modification for the calculated membership degrees. According to equations (3.7-3.12), the comprehensive reliability indexes for $C_j$ could be generated as $Rc_j$. Multiplied with the obtained $Rc_j$ based on (3.13), the original membership degree $M_{ijrl}$ can be modified as $M'_{ijrl}$. As there are $r$ computed modified membership degrees for each linguistic class, we aggregate them as follows:

$$M^*_{ijrl} = \sum_r M'_{ijl} \tag{5.2}$$

Then, the reliability modified membership degrees are normalized to induce TFN for the integrated TOPSIS decision matrix as mentioned below:

$$N_{ijl} = \frac{M^*_{ijl}}{\sum M^*_{ijl}}, \sum N_{ijl} = 1 \tag{5.3}$$

To generate the TFN for the attribute involving non-time series graphical information, we utilized the TFN for every linguistic class in Table 3 and the membership degrees calculated above. In this integration process, without loss of information generality, the membership degree is regarded as the weight of each linguistic class for every alternative concerning each attribute. Then the membership degree is converted to TFN that will be used in the TFN based TOPSIS decision matrix. The integrated TFN is presented as $A_{ij}$ $(a_{ij}, b_{ij}, c_{ij})$:

$$\begin{aligned} a_{ij} &= \sum_l a_l N_{ijl} \\ b_{ij} &= \sum_l b_l N_{ijl} \\ c_{ij} &= \sum_l c_l N_{ijl} \end{aligned} \tag{5.4}$$

Where $l$ is the linguistic class in Table 3.

***Step 2: Processing qualitative data***

Step 2 process and transfer qualitative data associated with supplier performance and importance weight of underlying attributes evaluation provided by multiple decision makers (DM). Following two sub-steps entails step 2.

***Step 2(a): Processing and transferring qualitative data entailing suppliers' performance evaluation to aggregated TFN***

The qualitative assessments given by the decision makers for each supplier against each attribute are directly transferred to respective TFNs. For example, a set of qualitative assessments for $i^{th}$ suppliers on attribute $C_j$ given by $k^{th}$ DM can be represented as in Table 4. The qualitative assessment given by $L_{ijk}$ can assume any of the form given by Bad (B), Moderately Bad (MB), Moderate (M), Moderately Good (MG), Good (G), Very Good (VG), Very Very Good (VVG), and Extremely Good (EG).

Table 4
Example of original linguistic data

| | | $C_j$ | | |
|---|---|---|---|---|
| $Supplier/DMs$ | $DM_1$ | $DM_2$ | $\dots$ | $DM_k$ |
| $S_1$ | $L_{1j1}$ | $L_{1j2}$ | $\dots$ | $L_{1jk}$ |
| $S_2$ | $L_{2j1}$ | $L_{2j2}$ | $\dots$ | $L_{2jk}$ |
| $\dots$ | $\dots$ | $\dots$ | $\dots$ | $\dots$ |
| $S_i$ | $L_{ij1}$ | $L_{ij2}$ | $\dots$ | $L_{ijk}$ |

These qualitative appraisals are then converted to respective TFNs using standard TFN (Table 14 in Appendix A) associated with different linguistic classes, similar to (C.-T. Chen et al., 2006). The converted TFN can be presented as in Table 5. In doing so, the uncertainty associated with vague qualitative assessment is also quantified with the help TFN.

Table 5
Example of TFN decision matrix

| | | $C_j$ | | |
|---|---|---|---|---|
| $Supplier/DMs$ | $DM_1$ | $DM_2$ | $\dots$ | $DM_k$ |
| $S_1$ | $(a_{1j1}, b_{1j1}, c_{1j1})$ | $(a_{1j2}, b_{1j2}, c_{1j2})$ | $\dots$ | $(a_{1jk}, b_{1jk}, c_{1jk})$ |
| $S_2$ | $(a_{2j1}, b_{2j1}, c_{2j1})$ | $(a_{2j2}, b_{2j2}, c_{2j2})$ | $\dots$ | $(a_{2jk}, b_{2jk}, c_{2jk})$ |
| $\dots$ | $\dots$ | $\dots$ | $\dots$ | $\dots$ |
| $S_i$ | $(a_{ij1}, b_{ij1}, c_{ij1})$ | $(a_{ij2}, b_{ij2}, c_{ij2})$ | $\dots$ | $(a_{ijk}, b_{ijk}, c_{ijk})$ |

Finally, the TFN-based TOPSIS decision matrix is constructed by incorporating and aggregating the conflicting qualitative assessments provided by all the DMs involved in the decision-making process. This is similar to what is proposed by C.-T. Chen et al. (2006) as follows:

$$A_{ij}\left(a_{ij}, b_{ij}, c_{ij}\right) = (\min_k a_{ijk}, \frac{\sum_k b_{ijk}}{k}, \max_k c_{ijk}) \tag{5.5}$$

***Step 2(b): Transferring qualitative data on attributes' weight to TFN***

The weights of all the attributes are determined and expressed by the DMs in the form of qualitative assessment as well. Such qualitative assessment can be expressed in the form of linguistics terms e.g., Very Unimportant (VUI), Unimportant (UI), Moderately Important (MI), Important (I), Very Important (VI), and Extremely Important (EI). Some of these linguistic terms and associated TFNs are listed in Table 6. Since the weight lies in between 0 to 1, the frame of discernment is represented as $U = [0,1]$. Using these TFNs we fuzzified the frame of discernment of the attribute weights as shown in Figure 10:

Table 6.
Linguistic terms and corresponding fuzzified TFN

| Weight of Criteria | |
|---|---|
| Linguistic Terms | TFN (a, b, c) |
| VUI | (0,0.1,0.2) |
| UI | (0.1,0.2,0.3) |
| … | … |
| EI | (0.8,0.9,1) |

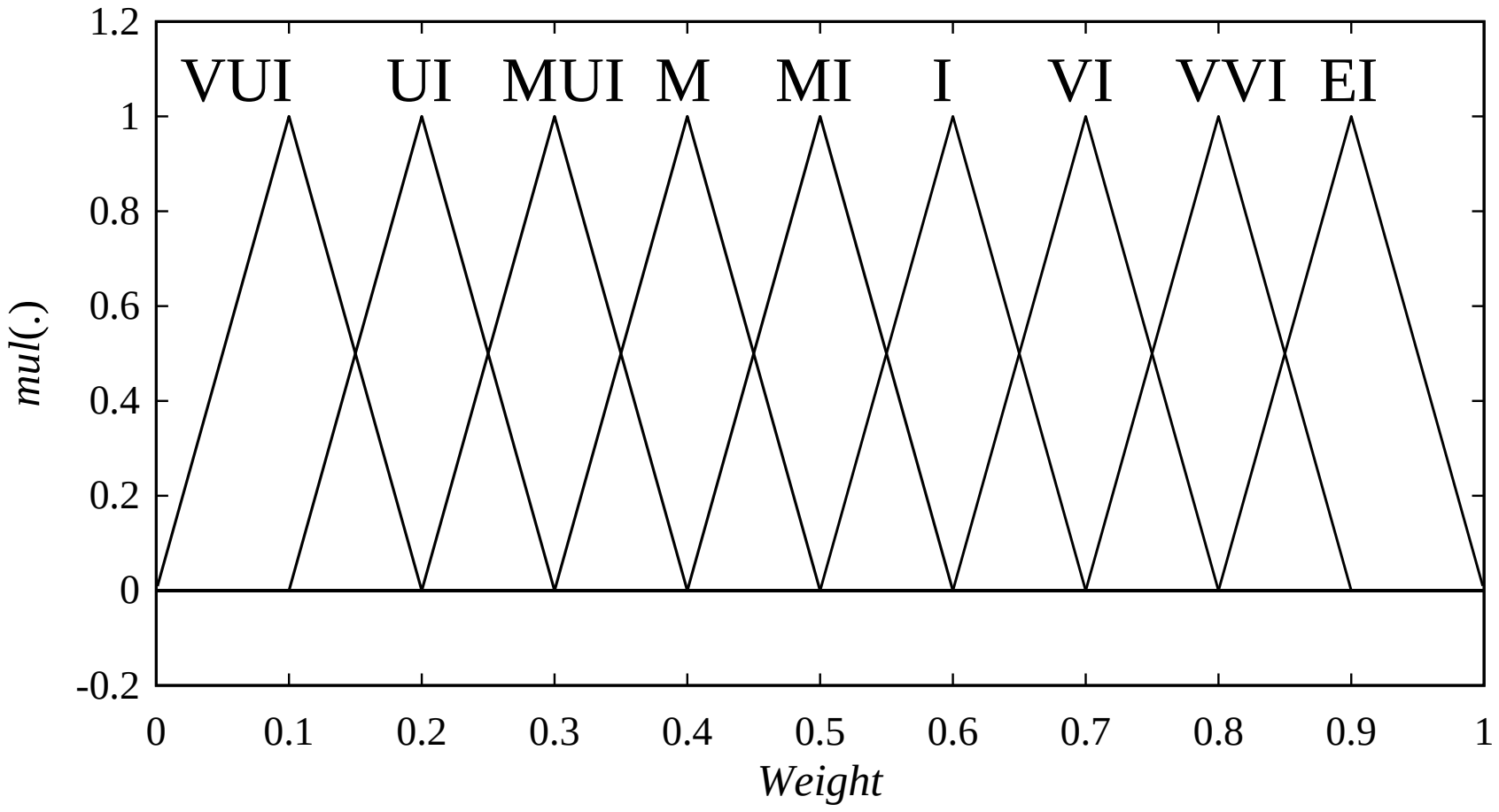

Figure 10. Fuzzification of criteria weight.

Then the weight of each attribute is at first directly converted to a TFN, $w_j(a_j, b_j, c_j)$ using the TFN associated with respective linguistic term. Once all the qualitative weights provided by multiple DMs are converted to respective TFNs, the aggregated weight and corresponding TFNs are generated according to equation 5.5 as mentioned in step 2(a).

***Step 3: Performing TOPSIS to rank alternative suppliers***

Because in the weighted decision matrix, the TFNs concerning each supplier against each attribute have different support defined as $[a_{ij,} c_{ij}]$,we first normalized each TFN $A_{ij}$ $(a_{ij}, b_{ij}, c_{ij})$ on all attributes before performing TOPSIS based on the principle used in (C.-T. Chen et al., 2006):

$$A'_{ij}\left(a'_{ij}, b'_{ij}, c'_{ij}\right) = \left(\frac{a_{ij}}{\max_i c_{ij}}, \frac{b_{ij}}{\max_i c_{ij}}, \frac{c_{ij}}{\max_i c_{ij}}\right), \forall\, j \in G_1 \quad (5.6)$$

$$A'_{ij}\left(a'_{ij}, b'_{ij}, c'_{ij}\right) = \left(\frac{\min_i a_{ij}}{c_{ij}}, \frac{\min_i a_{ij}}{b_{ij}}, \frac{\min_i a_{ij}}{a_{ij}}\right), \forall\, j \in G_2 \quad (5.7)$$

where $G_1$ is the set of beneficial attributes which will be maximized and $G_2$ is the set of non- beneficial attributes which will be minimized.

As now we have the normalized TFN $A'_{ij}\left(a'_{ij}, b'_{ij}, c'_{ij}\right)$ for all suppliers $S_i$ on every attribute $C_j$, and the attribute weight $w_j\left(a_j, b_j, c_j\right)$, the normalized and weighted TFN based TOPSIS decision matrix $\left\{A^*_{ij}\right\}$ is constructed based on (3.5):

$$A^*_{ij}(a^*_{ij}, b^*_{ij}, c^*_{ij}) = A'_{ij} \times w_j \quad (5.9)$$

The Positive Ideal Solution (PIS) and Negative Ideal Solution (NIS) are determined as (C.-T. Chen et al., 2006):

$$A_{pj}\left(a_{pj}, b_{pj}, c_{pj}\right) = \max_i c^*_{ij}$$
$$A_{nj}\left(a_{nj}, b_{nj}, c_{nj}\right) = \min_i a^*_{ij} \quad (5.10)$$

And, finally the closeness coefficient $(\tilde{\rho}_i)$ for $S_i$ is generated as:

$$d_i^+ = \sqrt{\frac{\sum_j (a^*_{ij} - a_{pj})^2 + \sum_j (b^*_{ij} - b_{pj})^2 + \sum_j (c^*_{ij} - c_{pj})^2}{3}}$$
$$d_i^- = \sqrt{\frac{\sum_j \left(a^*_{ij} - a_{nj}\right)^2 + \sum_j \left(b^*_{ij} - b_{nj}\right)^2 + \sum_j \left(c^*_{ij} - c_{nj}\right)^2}{3}} \quad (5.11)$$
$$\tilde{\rho}_i = \frac{d_i^-}{(d_i^- + d_i^+)}$$

The higher the $\tilde{\rho}_i$, the higher will be the ranking for a particular supplier. Ranking is given as an ascending order starting from 1 for a supplier with highest $\tilde{\rho}_i$, and follows chronological order for rest of the suppliers.

***Step 4: Performing MCGP to determine optimal order allocation policy***

It is perhaps not surprising that a single supplier may not always have the ability to supply the entire ordered quantity. In addition, strategic decision makers may opt for diversifying the sourcing channels while ensuring stability and competitiveness among alternative suppliers under the threat of disruption risks. These altogether make it feasible that decision makers often times have to depend on multiple suppliers, requiring an optimal order allocation strategy that takes into account the supplier preferential ranking $R_i$ generated via F-MADM approach as input. Multi-choice Goal Programming (MCGP)—a viable approach in this regard—can successfully be integrated with F-MADM based DSS to devise an optimal order allocation plan (Liao & Kao, 2011). MCGP has the potential to address multi-attribute decision making problems, wherein decision makers aim to minimize the penalty of a set of objectives assigned to all attributes. The essential idea of integrating F-MADM approach with MCGP is to enable such an optimal order allocation policy that can maximize the total value created from the intended procurement plan. The Total Value of Procurement (TVP) is quantitatively defined as the aspiration level that is set by the decision makers, and it can be sometimes conservative based on the company's resource limitations and incompleteness of available information. A well-judged and/or conservative aspiration level e.g. TVP can avoid the potential negative effect of the intended procurement plan. Often times, in MCGP setting, the decision makers are allowed to set a multi-choice aspiration level (MCAL) for each goal to help avoid unintended underestimation and overestimation of decision making (Chang, 2008). The MCAL for each target associated with multiple attributes is presented in certain interval values, allowing the decision makers to consider uncertainty/incompleteness of available decision relevant information. Therefore, once the closeness coefficients for each alternative suppliers are generated at the end of Step 3, these are used as the coefficient in the proposed MCGP according to (Guneri, Yucel, & Ayyildiz, 2009) so that the overall penalty for not satisfying the targets is minimized. .

Leveraging this principle, we formulated the MCGP as follows:

$$Minimize \sum_{i}(d_i^+ + d_i^-) + \sum_{j}(e_j^+ + e_j^-), \quad i = 1,2,3,4, j = 1,2,3$$

*Subject to:*

$$\sum_{1}^{n} C_n \times x_n - d_1^+ + d_1^- \geq T \qquad (1)$$

$$\sum_{1}^{n} U_n \times x_n - d_2^+ + d_2^- = y_1 \qquad (2)$$

$$y_1 - e_1^+ + e_1^- = I_{min} \qquad (3)$$

$$I_{min} \leq y_1 \leq I_{max} \qquad (4)$$

$$(\sum_{1}^{n} L_n \times x_n)/\sum_{1}^{n} x_n - d_3^+ + d_3^- = y_2 \qquad (5)$$

$$y_2 - e_2^+ + e_2^- = R_{min} \qquad (6)$$

$$R_{min} \leq y_2 \leq R_{max} \qquad (7)$$

$$\sum_{1}^{n} x_n - d_4^+ + d_4^- \leq Q \qquad (8)$$

$$x_n,\ d_i^+, d_i^-, e_j^+, e_j^- \geq 0 \qquad (9)$$

Where:

$d_i^+, d_i^-, e_m^+, e_m^-$ stand for the penalties in violation of respective constraints

$x_n$ is the optimal ordered quantity assigned to $n_{th}$ Supplier

$C_n$ is the closeness coefficients $(\tilde{\rho}_i)$ of the available suppliers

$T$ is the total value created from procurement (TVP)

$U_n$ is the unit cost of quantity when purchased from $n_{th}$ supplier

$y_1$ is the total available budget for the procurment

$I_{min},\ I_{max}$ $are\ the\ lower\ limit\ and\ upper\ limit\ on\ the\ budget, respectively$

$L_n$ is lead time of the $n_{th}$ supplier

$y_2$ is the total allowable lead time for a particualr order

$R_{min},\ R_{max}$ $are\ the$ lower limit and upper limit on lead time, respectively

$Q$ $is$ the procurement level set by the decision makers

$n$ is the number of alternative suppliers

The objective function aims to minimize the total non-achievement penalties of multiple targets assigned in different constraints. Constraint (1) ensures that the orders should be allocated among multiple suppliers considering their preferential ranking in such a way so that a minimum TVP is achieved. In other words, constraint (1) sets an upper bound on TVP. Meanwhile, constraint (2) refers to the goal of procurement budget, signifying that total procurement cost will not exceed the budget after including the positive and negative deviation of the intended goal. Constraints (3) and (4) explains the aspiration levels of the goal associated with procurement budget. In a similarly way, constraint (5), (6) and (7) illustrate the lead time preference along with the aspiration levels associated with corresponding lead times. Finally, constraint (8) with consideration of deviations from the procurement level goal, restrict that the total order allocated to multiple suppliers must equal the procurement level set by the decision makers.

Such an MCGP model is anticipated to handle multiple objectives if a decision maker seeks the optimal solution from a set of feasible solutions considering the aspiration levels of the objectives; thus, enabling the management to optimally balance their requirements among alternative suppliers when the multiple requirements cannot be satisfied by a single supplier.

The proposed decision-making framework is presented as a flow chart in Figure 11:

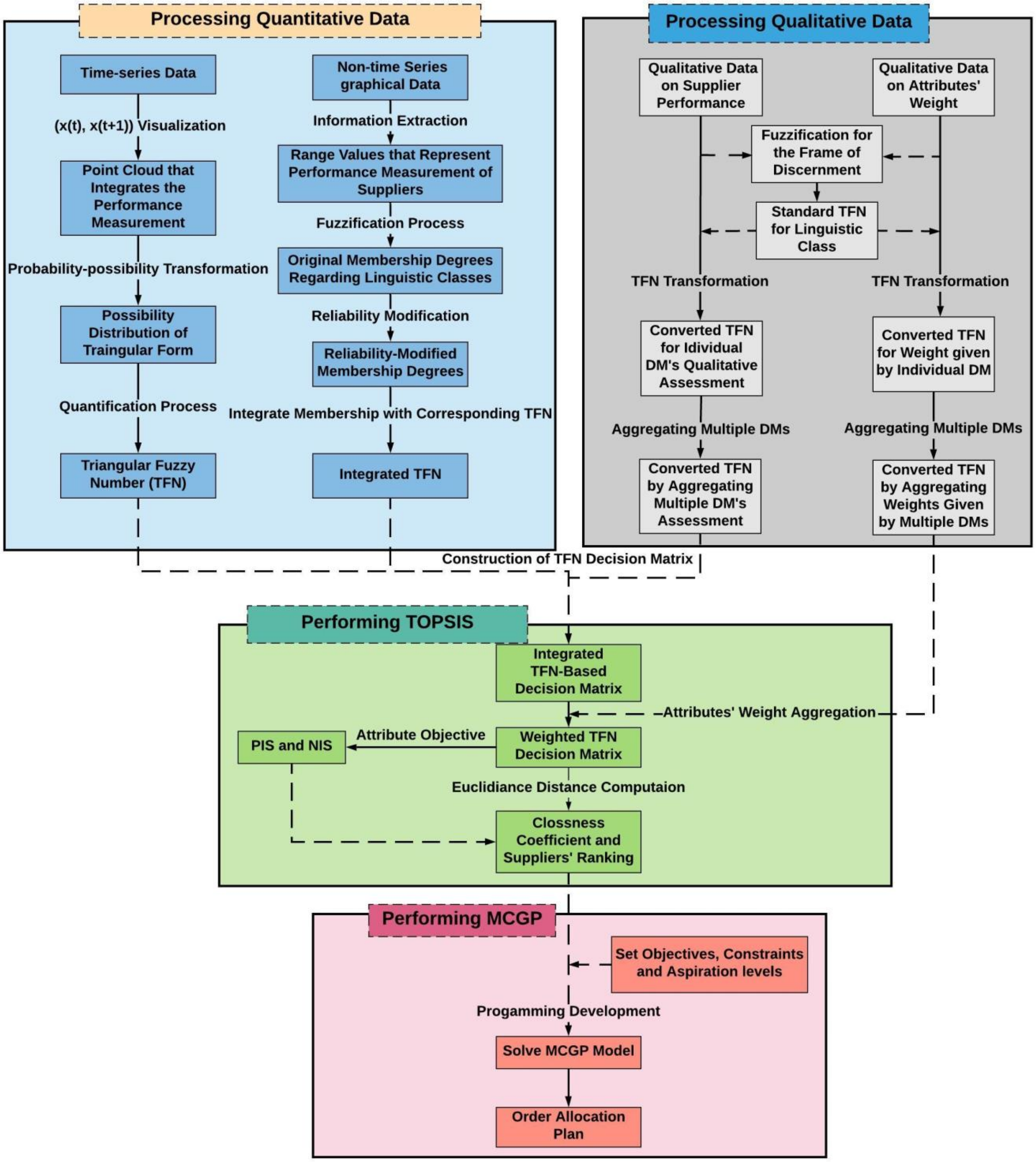

Figure 11. DSS for resilient supplier evaluation framework in logistics 4.0.

## 6. Case illustration

To demonstrate the effectiveness and usefulness of the proposed supplier evaluation and order allocation DSS, we present a hypothetical case study that is generalizable to companies operating under logistics 4.0. Such logistics 4.0 companies concern different aspects of end-to-end logistics and supply chain management, which transforms the way those companies manage their logistics operations. This transformation is powered by the digitalization of supply chain system—characterized by the speed, flexibility, real-time connectedness among different entities of logistics 4.0—arguably, supply chain 4.0. Crucially, effective sourcing of raw materials plays a significant role in achieving that desired level of efficiency, responsiveness and resilience in the context of connected, decentralized and digitalized supply chain. Often times, a set of alternative suppliers may serve the purpose of providing a particular raw material. Strategic decision makers responsible for taking such high-impact sourcing decisions must choose a supplier among available alternatives who can best serve the requirement of resilience, sustainability and efficiency. Therefore, the supplier evaluation and selection process in this context of logistics 4.0 is characterized as a decision-making problem comprising multiple conflicting attributes. The problem becomes even more complex when a single supplier is not able to provide entire ordered quantity, and allocation of order is needed among multiple suppliers.

### 6.1 Evaluation attributes

As alluded previously, we select and define several attributes based on which the alternative suppliers are evaluated from logistics 4.0 and resilience perspectives. Attributes are divided into two main groups: (1) quantitative and (2) qualitative as presented in Table 7. For attributes in the quantitative subset, large number of data is available in the form of continuous time series. Data collected historically are characterized as non-time series data. These historically collected data are often stored graphically in logistics 4.0 environment due to the digitalization, cloud storage facilities and Internet of Things (IOT). Moreover, when visualized graphically, these data entail a great deal of actionable information that has been proven to be valuable while evaluating several sourcing options. As such, quantitative criteria are then sub-divided into two groups based on the type of data available as decision relevant information. We characterized that for inventory and delivery schedules, data are collected continuously and presented in time series as real-time visibility and end-to-end data sharing are considered as crucial aspects of logistics 4.0. On the other hand, data associated with supplier's production capacity and cost are collected over the time and can be presented graphically. In case of other fifteen attributes listed in Table 7, qualitative assessments are given by multiple decision makers for each alternative supplier. All those attributes are so chosen that has been used to measure the resilience performance of the suppliers in the context of logistics 4.0.

To further specify the effect of these attributes in enhancing resilience i.e., reducing vulnerability against anticipated disruptions and improving recoverability after being affected by disruption, we categorize and associate them to pre-disaster and post-disaster resilience activities. Attributes $C_2$, $C_5$, $C_6$, $C_7$, $C_8$, $C_9$, $C_{11}$, $C_{12}$, $C_{13}$, and $C_{15}$ are used to evaluate alternative suppliers based on their ability to reduce the vulnerability against potential disruptions, and thus refer to the pre-disaster resilience activities of the suppliers. On the other hand, attributes $C_1$, $C_3$, $C_{10}$, $C_{14}$, $C_{16}$, $C_{17}$, $C_{18}$ and $C_{19}$ are used to evaluate suppliers depending on their ability to recover quickly and effectively after being affected by disruption, and thus represent supplier's post disaster resilience strategies. Cost (attribute $C_4$) is considered as expense that supplier has to incur to provide the goods, and also to ensure the desired level of resilience through coordinated pre-disaster and post-disaster strategies.

Table 7.
List of attributes considered in decision making process

| | Types of decision relevant information | $C_j$ | Attributes | Objective | Description |
|---|---|---|---|---|---|
| Quantitative attributes | Time-series data | $C_1$ | Pre-positioned inventory level | Max | The quantity of inventory in stock and available for supply. |
| | | $C_2$ | Lead time variability | Min | Time that supplier take to deliver the order to the company. |
| | Non-time-series data presented graphically | $C_3$ | Production capacity | Max | Quantity of the products that a supplier is capable to produce per day. |
| | | $C_4$ | Cost | Min | Cost that is incurred by the company while purchasing the required quantity from a particular supplier. The cost here included per unit production and transportation cost. |
| Qualitative attributes | Qualitative assessment presented in linguistic terms | $C_5$ | Digitalization | Max | Enabled by Web technologies, work flow tools, portals for customers, suppliers and employees, and information technology innovations targeted at supply chains and customer relationships (Rai et al., 2006). |
| | | $C_6$ | Traceability | Max | The ability to trace the origin of materials and parts, processing history and distribution or |

| | | | | |
|---|---|---|---|---|
| | | | | location of the product while being delivered (Aung & Chang, 2014). |
| | $C_7$ | Supply chain density | Min | The quantity and geographical spacing of nodes within a supply chain. |
| | $C_8$ | Supply chain complexity | Max | The number of nodes in a supply chain and the interconnections between those nodes. |
| | $C_9$ | Re-engineering | Max | The corrective procedure for the incorporation of any engineering design change within the product. Suppliers need to possess re-engineering capability to respond to customer's change of taste or requirements. |
| | $C_{10}$ | Supplier's resource flexibility | Max | The different logistics strategies which can be adopted either to release a product to a market or to procure a component from a supplier. |
| | $C_{11}$ | Automation disruption | Min | Ability to withstand the disruption caused in the automated manufacturing system. |
| | $C_{12}$ | Information management | Max | The ability to acquire, store, retrieve, process and share fast flowing information regarding demand and lead time volatility, change in price, real time location sharing while delivering the raw materials. |
| | $C_{13}$ | Cyber security risk management | Max | Ability to prevent or mitigate damage from IT security breaches in supply chains, where breaches can disrupt production, cause loss of essential data, and compromise confidential information. |
| | $C_{14}$ | Supplier reliability | Max | The availability during disruptions of alternative transportation channels with different characteristics based on their costs and delivery dates. |
| | $C_{15}$ | Supply chain visibility | Max | The ability of the supplier to have a vivid view of upstream and downstream inventories, |

| | | | |
|---|---|---|---|
| | | | demand and supply conditions, and production and purchasing schedules. |
| $C_{16}$ | Level of collaboration | Max | Supplier collaboration reduces forecasting and inventory management risks, thereby enhancing resilience of supply chains. Also, it helps mitigate supply side uncertainty after disruption hits. |
| $C_{17}$ | Restorative capacity | Max | The ability of suppliers to repair and quickly restore to its normal operating conditions after a disruptive event. |
| $C_{18}$ | Rerouting | Max | Capability of changing the usual mode of transport while anticipating or being affected by the disruptions. Companies can combine multiple modes of intermodal transportation which are fast to ensure uninterrupted supply of goods and operations of supply chain. |
| $C_{19}$ | Agility | Max | The speed with which a firm's internal supply chain functions can adapt to marketplace changes resulting from disruption and thus can better respond to unforeseen events |

### 6.2 Results and sensitivity analysis

To test the practicability of our proposed model, we randomly generated a set of data in Appendix A. The numerical example includes five alternative suppliers that are evaluated with regards to four quantitative attributes and fifteen qualitative attributes presented in Table 7. For each supplier, 500 records are collected as continuous time series in case of pre-positioned inventory level (attribute $C_1$) and lead time variability (attribute $C_2$) (presented in Figure 15 & Figure 16 in Appendix A). As mentioned in sub-step 1(a) in section 5, these time-series data are transferred to possibility distribution of triangular form (Figure 18 & Figure 19 in Appendix B), which afterwards were used to induce TFNs (Table 15 in Appendix B). In case of production capacity and cost, 300 records are used from historically collected data, which are presented in several pieces of graphs, making it easier to process this large quantity of data into actionable decision relevant information (Figure 17 in Appendix A). According to the sub-step 1(b) mentioned in section 5, the crisp granular information extracted from these graphs concerning each supplier are presented in Table 16 (Appendix B). Then the integrated TFNs associated with each of these two attributes for all five alternative suppliers are computed following the procedure described in sub-step 1(b) in section 5 and are presented in Table 17 (Appendix B). Performance evaluation data concerning each supplier against each qualitative attribute are collected from five decision makers who are assumed to have equal importance in decision making process. As previously mentioned in section 5, these qualitative assessments are provided in the form of linguistic appraisals, which are presented in Table 13 for attribute $C_5$ to $C_{19}$ (Appendix A). These qualitative assessments are transferred to corresponding TFNs according to the process detailed in sub-step 2(a) in section 5. Similarly, the qualitative weights of all the attributes provided by multiple DMs in linguistic terms are also listed in Table 13 (Appendix A), which later are converted to respective TFNs according to the principle described in sub-step 2(b) in section 5 and presented in Table 18 (Appendix B). After converting all the quantitative and qualitative DRI into respective TFNs, the weighted TFN-based TOPSIS decision matrix is constructed, which was used to determine the PIS and NIS according to step 3 in section 5. The PIS and NIS associated with all the attributes are listed in Table 19 (Appendix B). Finally, the closeness coefficients ($\tilde{\rho}_i$) are calculated for each alternative supplier.

The closeness coefficient ($\tilde{\rho}_i$) and preferential ranking scores of the alternative suppliers generated by the proposed model are presented in the Table 8 and Figure 12. As presented in Figure 12, supplier 3 has the highest $\tilde{\rho}_i$ value, whereas lowest $\tilde{\rho}_i$ value is observed for supplier 5. The highest $\tilde{\rho}_i$ value indicates highest preferential ranking for that particular supplier, and corresponds to lowest number in the $R_i$. The lowest number in $R_i$ stands for highest preferred supplier.

Table 8.
Ranking score of the suppliers

| Supplier | $d^+$ | $d^-$ | $\tilde{\rho}_i$ | $R_i$ |
|---|---|---|---|---|
| $S_1$ | 1.80 | 1.39 | 0.436 | 3 |
| $S_2$ | 1.82 | 1.44 | 0.441 | 2 |
| $S_3$ | 1.76 | 1.48 | 0.456 | 1 |
| $S_4$ | 1.88 | 1.33 | 0.414 | 4 |
| $S_5$ | 1.95 | 1.23 | 0.388 | 5 |

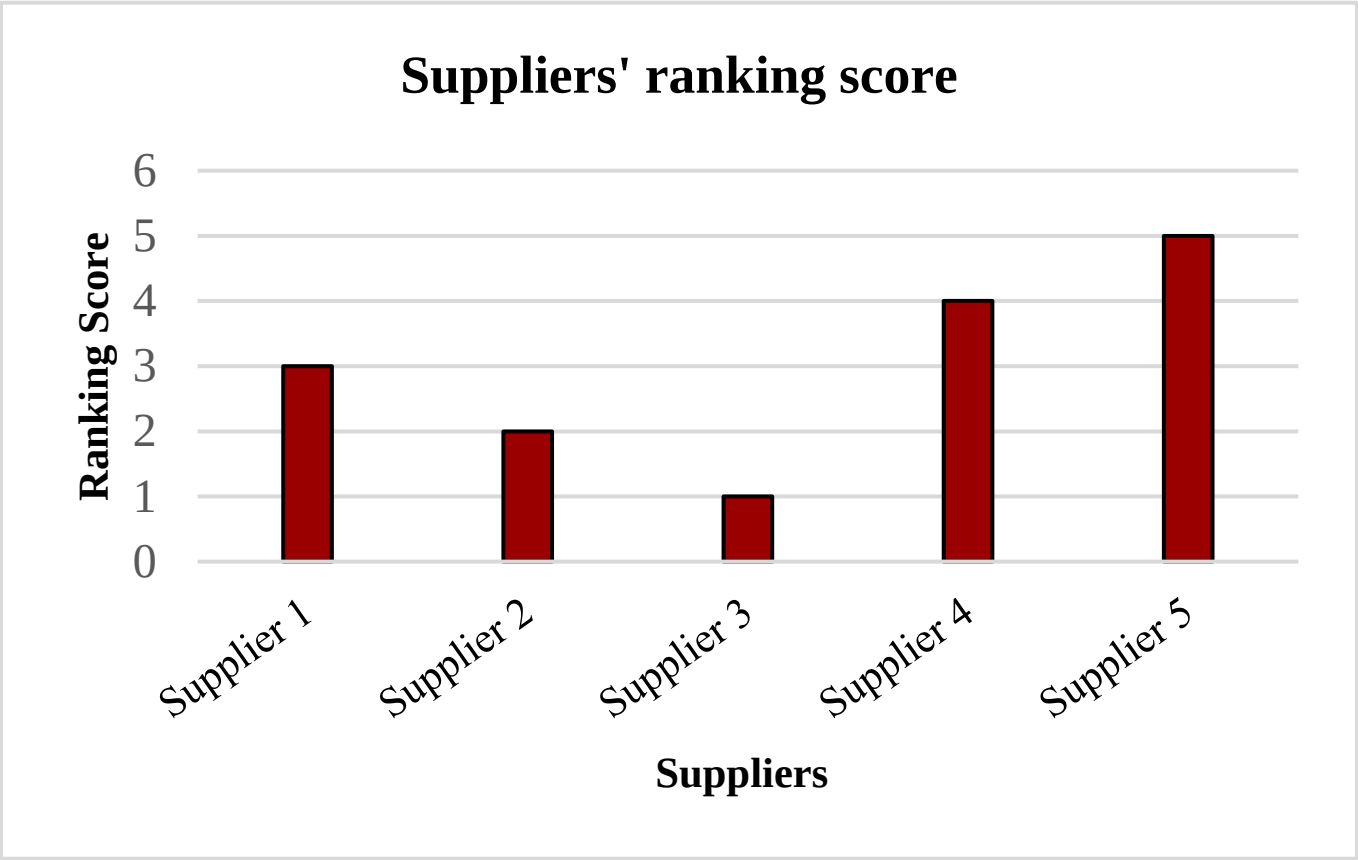

Figure 12. Supplier's ranking score

Once the $\tilde{\rho}_i$ values are obtained for each supplier, the MCGP is performed and solved with Lingo software. We assume that based on the previous experience, available resources and information, the decision makers have set different aspiration levels associated with different goals, which are presented below:

1) The total value created from procurement, defined as $T$ at least 260, and the more the better. It is signified as TVP constraint.
2) The total cost of procurement is set in between \$300000 to \$350000, and the less the better. This correspond to budget constraint.
3) The lead time alternatively termed as delivery time is set in between 10 and 12 days, and the less the better. These parameters are used for delivery constraint.
4) Total procurement level should not be higher than 500, leads to order quantity constraint.

With the abovementioned multiple goals, the MCGP is formulated as follows:

$$Minimize \sum_i (d_i^+ + d_i^-) + \sum_j (e_j^+ + e_j^-), \quad i = 1,2,3,4, j = 1,2,3$$

*Subject to:*

$$0.467x_1 + 0.45x_2 + 0.448x_3 + 0.451x_4 + 0.388x_5 - d_1^+ + d_1^- \geq 260 \qquad (1)\ ; TVP\ constraint$$

$$700x_1 + 1000x_2 + 600x_3 + 500x_4 + 650x_5 - d_2^+ + d_2^- = y_1 \qquad (2)\ ; Budget\ constraint$$

$$y_1 - e_1^+ + e_1^- = 300000 \qquad (3)\ ; Budget\ aspiration\ level$$

$$250000 \leq y_1 \leq 350000 \qquad (4)\ ; Budget\ aspiration\ level$$

$$(11.01x_1 + 9x_2 + 14.03x_3 + 14.01x_4 + 14x_5) \Big/ \sum_1^n x_n - d_3^+ + d_3^- = y_2 \qquad (5)\ ; Delivery\ time\ constraint$$

$$y_2 - e_2^+ + e_2^- = 10 \qquad (6)\ ; Delivery\ time\ aspiration\ level$$

$$10 \le y_2 \le 12 \qquad (7)\ ;\textit{Delivery time aspiration level}$$

$$\sum_{1}^{n} x_n - d_4^+ + d_4^- \le 500 \quad (8)\ ;\textit{Order quantity constraint}$$

$$x_n,\ d_i^+, d_i^-, e_j^+, e_j^- \ge 0 \quad (9)$$

After solving the formulated MCGP model, the results generated from Lingo are presented in Table 9:

Table 9.
Optimal order allocation plan

| Supplier | Allocated Order Quantity |
|---|---|
| $S_1$ | 29 |
| $S_2$ | 0 |
| $S_3$ | 442 |
| $S_4$ | 29 |
| $S_5$ | 0 |

Thus, in the final order allocation plan, the order quantity assigned to $S_1$, $S_3$ and $S_4$ are 29, 442 and 29 respectively with the total order quantity of 500, while other suppliers are not assigned with any order quantity. Additionally, it is perhaps not surprising that based on the company's available resources and information concerning the alternative suppliers, management may set different aspiration level for TVP goal ranging from most pessimistic to most optimistic estimation. Thus, the DSS system should be able to propose alternative order allocation plan subject to the change of aspiration level associated with TVP goal. Therefore, we investigate several other instances by changing the aspiration level of TVP and assessed the effect of different TVP value on the order allocation plan as presented in Figure 13.

For a more pessimistic estimation of TVP within 160 to 180, the model assigns order to supplier 1 and supplier 2, with higher preference given to the first supplier. The allocated order quantity to supplier 1 increases up until a TVP value of 190, beyond which it starts to decrease and gets stable at a TVP value of 230. Orders are allocated to supplier 4 at a TVP value of 190, increases up until a TVP of 210, and similar to supplier 1 gets stable at TVP of 230. At a higher TVP value e.g., 220, majority of the order is allocated to highest ranked supplier 3 with equal quantity of order allocated to supplier 1 and supplier 4. After TVP value of 230, a stable order allocation plan is achieved entailing supplier 3, supplier 1 and supplier 4 with highest priority given to supplier 3.

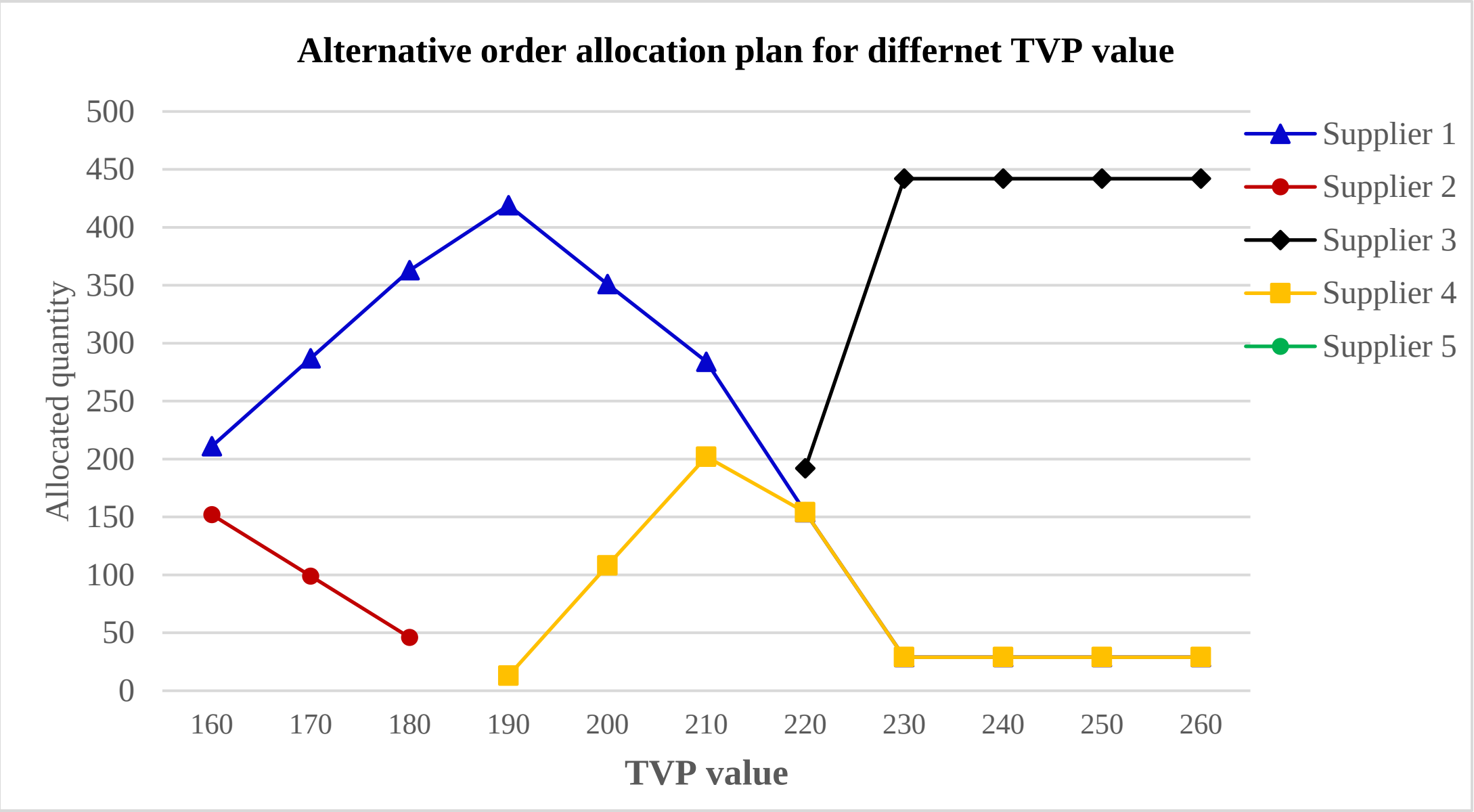

Figure 13. Alternative order allocation plan for different TVP value

The preferential ranking score generated from F-MADM approach considers refined weights provided by the decision makers on attributes involved in alternative supplier selection process. This is essential as not all attributes are equally important in supplier evaluation scheme, and thus for making rational decisions, the differential weights are incorporated in our proposed F-MADM based DSS. However, often times cost factor associated with a procurement plan—and largely with a supplier, can turn out to be a vital attribute, requiring greater importance on cost over all other decision attributes. It is especially true for a logistics 4.0 company that mostly prefers efficiency from a supplier. On the contrary, prioritizing resilience performance of a supplier over cost may often be the dominating preference for type of logistics 4.0 company valuing greater ability against disruption, and thus enhancing visibility and responsiveness in satisfying customer needs even at the expense of greater cost. Reflecting on these two extremely opposite needs of the management, we investigate how to assist strategic decision makers in analyzing this trade-off while evaluating alternative suppliers for sourcing options. In what follows, we categorize the attributes mentioned in section 6.1 into two segments—considering cost alone as measure of efficiency while the rest of the attributes as-a-whole are considered as a holistic measure of resiliency for a supplier. We then perform TOPSIS separately on these two sets of attributes. Precisely saying, the $\tilde{\rho}_i$ generated in the context of resiliency will not include the suppliers' information on cost attribute, which means in 5.11, $j \neq C_4 \; \forall \; \tilde{\rho}_{iR}$ and $j = C_4 \; \forall \; \tilde{\rho}_{iC}$; $\tilde{\rho}_{iR}$ and $\tilde{\rho}_{iC}$ refer to the closeness coefficient associated with the resiliency and efficiency measure, respectively for $i^{th}$ supplier.

It is anticipated that supplier's preferential ranking will change based on the differential importance on efficiency and resilience measures. To categorically distinguish this from the originally generated ranking score by F-MADM, we use Supplier's Cost versus Resilience Index (SCRI) as a trade-off measure between resilience and efficiency, which is defined as follows:

$$SCRI_i = \alpha\,\tilde{\rho}_{iR} + (1-\alpha)\tilde{\rho}_{iC} \qquad (6.1)$$

Where $SCRI_i$ is the index of $i^{th}$ supplier, and $\alpha$ refers to the importance in the range of [0, 1] given by the decision makers on resiliency performance. The lowest $\alpha$ value is explained as decision maker's lowest importance on resilience measure and vice-versa for cost (efficiency) measure. The corresponding $\tilde{\rho}_i$ values (also normalized) as measured by TOPSIS for cost attributes and resilience attributes are presented in Table 10 and Table 11:

Table 10
Closeness Coefficient ($\tilde{\rho}_i$) for resilience attributes

| | | $\tilde{\rho}_{iR}$ | | |
|---|---|---|---|---|
| Supplier | d+ | d- | $\tilde{\rho}_i$ | Normalized |
| $S_1$ | 1.78 | 1.38 | 0.436 | 0.2043 |
| $S_2$ | 1.80 | 1.43 | 0.442 | 0.2064 |
| $S_3$ | 1.74 | 1.46 | 0.457 | 0.2137 |
| $S_4$ | 1.86 | 1.30 | 0.413 | 0.1939 |
| $S_5$ | 1.93 | 1.21 | 0.387 | 0.1817 |

Table 11.
Closeness Coefficient ($\tilde{\rho}_i$) for cost (efficiency) attribute

| | | $\tilde{\rho}_{iC}$ | | |
|---|---|---|---|---|
| Supplier | d+ | d- | $\tilde{\rho}_i$ | Normalized |
| $S_1$ | 0.45 | 0.43 | 0.49 | 0.2073 |
| $S_2$ | 0.50 | 0.36 | 0.42 | 0.1762 |
| $S_3$ | 0.48 | 0.39 | 0.45 | 0.1913 |
| $S_4$ | 0.44 | 0.45 | 0.51 | 0.2143 |
| $S_5$ | 0.45 | 0.44 | 0.50 | 0.2110 |

Using normalized $\tilde{\rho}_{iR}$, $\tilde{\rho}_{iC}$ and $\alpha$ values, we then investigate the change of SCRI for different suppliers as presented in Figure 14. Supplier 4 has the highest SCRI value till $\alpha = 0.4$, pointing to the fact that when seeking resilience is less important compared to efficiency (signified by lower $\alpha$ values), supplier 4 is highly preferred being the most efficient or cost-effective supplier. Within a range of $\alpha$ in between 0.4 to 0.61, supplier 1 has the highest SCRI value. It suggests that when the importance of being efficient and resilient is almost equal or does not differ that much, supplier 1 should be preferred. After a value of $\alpha$ = 0.61, supplier 3 has shown highest SCRI, indicating that when higher preference is given on resiliency, supplier 3 dominates all other alternative suppliers. Although supplier 5 has relatively higher SCRI value when higher importance is given on efficiency measure, its SCRI value decreases with higher importance given on resilience measure. On the other hand, for supplier 2 and supplier 3, the SCRI values increase with higher importance given on resilience measure. For several combinations of resilience versus efficiency trade-off, the generated SCRI values are presented in the Table 12 in Appendix A. Thus, our proposed DSS has demonstrated the managerial implication in terms of assisting the strategic decision makers to analyze the resiliency versus efficiency trade-off while evaluating and selecting alternative suppliers along with the corresponding order allocation plan.

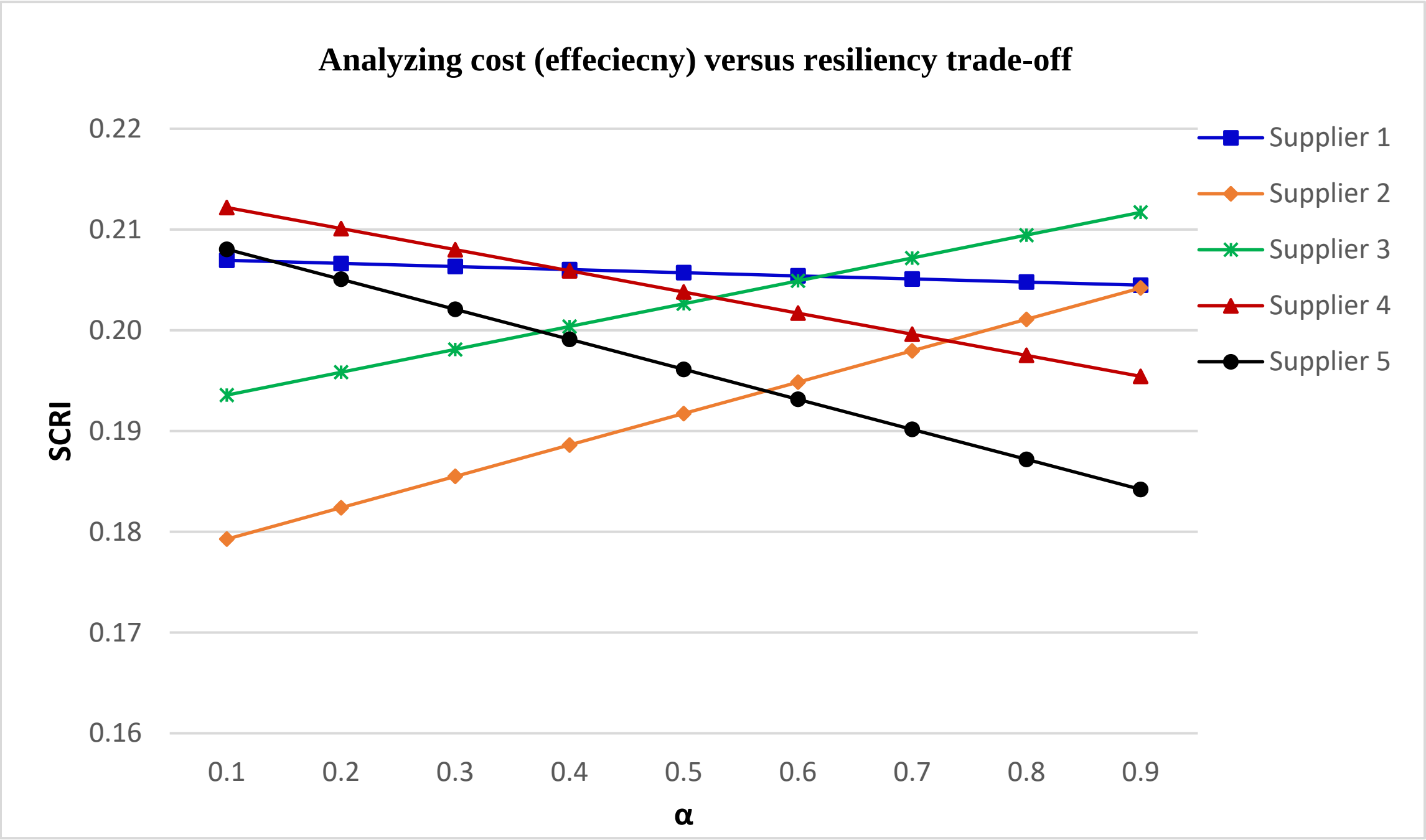

Figure 14. Sensitivity analysis by assessing resiliency versus efficiency trade-off

## 7. Conclusion

Selection of resilient suppliers in the context of logistics 4.0 requires processing heterogeneous information originated from multiple qualitative and quantitative attributes that are conflicting in nature. Additionally, most of the qualitative attributes considered to measure the performance of resilient suppliers in logistics 4.0 are substantially different than those used in traditional supplier selection problem—a combined fact that limits the applicability of traditional Fuzzy-Based Supplier selection framework in the presence of heterogeneous DRI. To address these issues, this paper presents a DSS that considers the inherent uncertainty of imprecise DRI to rank a set of alternative suppliers from resilient and logistics 4.0 point of view. Particularly, we adapted and extended the TFN based TOPSIS to the framework of logistics 4.0 that can handle qualitative information and large number of quantitative information presented in the time-series as well as graphical format. Using static and dynamic reliability index, we modified the membership value to further take into account the uncertainty and impreciseness of triangular membership function. Because one supplier may sometimes fail to provide the entire ordered quantity, we develop a model leveraging MCGP to allocate order among alternative supplies. This model takes input from the supplier ranking scores generated by proposed F-MADM approach. We also investigate the sensitivity of supplier's resiliency versus efficiency measures with the change in importance of resiliency attributes (from resilience and logistics 4.0 perspective) and cost attribute. That way, we empower the decision makers to generate alternative index based on the differential importance on resiliency and cost attributes. We believe, the developed DSS will provide an effective and pragmatic approach to help stakeholders devise better sourcing decisions for logistics 4.0 industries. Future research can explore how to incorporate interdependencies among several attributes that often times possess hierarchically structured relationship to some extent. Further research can also be carried out to explore other techniques such as PROMETHEE along with other fuzzy sets such as Interval Valued Intuitionistic Fuzzy Sets (IVFS) to rank alternative suppliers supported by a mechanism for optimizing weights of the associated decision makers. In addition to that, future studies can consider new attributes in MADM framework or adding constraint in the MCGP model to adapt with any policy changes within the company due to the disruptions.

**Funding**

This research did not receive any specific grant from funding agencies in the public, commercial, or not-for-profit sectors.

## Reference

Ashtiani, B., Haghighirad, F., Makui, A., & ali Montazer, G. (2009). Extension of fuzzy TOPSIS method based on interval-valued fuzzy sets. *Applied Soft Computing, 9*(2), 457-461.

Atanassov, K., & Gargov, G. (1989). Interval valued intuitionistic fuzzy sets. *Fuzzy Sets and Systems, 31*(3), 343-349. doi: 10.1016/0165-0114(89)90205-4

Atanassov, K. T. (1999). Intuitionistic fuzzy sets *Intuitionistic fuzzy sets* (pp. 1-137): Springer.

Aung, M. M., & Chang, Y. S. J. F. c. (2014). Traceability in a food supply chain: Safety and quality perspectives. *39*, 172-184.

Barreto, L., Amaral, A., & Pereira, T. J. P. M. (2017). Industry 4.0 implications in logistics: an overview. *13*, 1245-1252.

Bhutia, P. W., & Phipon, R. J. I. J. o. E. (2012). Application of AHP and TOPSIS method for supplier selection problem. *2*(10), 43-50.

Bohlender, G., Kaufmann, A., & Gupta, M. M. (1986). *Introduction to fuzzy arithmetic*.

Boran, F. E., Genç, S., Kurt, M., & Akay, D. (2009). A multi-criteria intuitionistic fuzzy group decision making for supplier selection with TOPSIS method. *Expert Systems with Applications, 36*(8), 11363-11368.

Brettel, M., Friederichsen, N., Keller, M., & Rosenberg, M. (2014). How virtualization, decentralization and network building change the manufacturing landscape: An Industry 4.0 Perspective. *International journal of mechanical, industrial science and engineering, 8*(1), 37-44.

Chang, C.-T. (2008). Revised multi-choice goal programming. *Applied mathematical modelling, 32*(12), 2587-2595.

Chen, C.-T., Lin, C.-T., & Huang, S.-F. (2006). A fuzzy approach for supplier evaluation and selection in supply chain management. *International Journal of Production Economics, 102*(2), 289-301. doi: 10.1016/j.ijpe.2005.03.009

Chen, T.-Y. (2015). The inclusion-based TOPSIS method with interval-valued intuitionistic fuzzy sets for multiple criteria group decision making. *Applied Soft Computing Journal, 26*, 57-73. doi: 10.1016/j.asoc.2014.09.015

Chen, T.-Y., Wang, H.-P., & Lu, Y.-Y. (2011). A multicriteria group decision-making approach based on interval-valued intuitionistic fuzzy sets: A comparative perspective. *Expert Systems with Applications, 38*(6), 7647-7658. doi: 10.1016/j.eswa.2010.12.096

Corporation, R. E. (2015). Renesas Electronics Enables Increased Factory Productivity for Industry 4.0 with New Industrial Ethernet Communication IC with Built-In Gigabit PHY. Retrieved 02/26, 2019, from https://www.renesas.com/us/en/about/press-center/news/2015/news20150625.html

De Felice, F., Deldoost, M. H., & Faizollahi, M. J. I. J. o. E. B. M. (2015). Performance measurement model for the supplier selection based on AHP. *7*(Godište 2015), 7-17.

Dempster, A. P. (1967). Upper and Lower Probabilities Induced by a Multivalued Mapping. *The Annals of Mathematical Statistics, 38*(2), 325-339. doi: 10.1214/aoms/1177698950

Devlin, N., Sussex, J. J. M., & Economics, p. L. O. o. H. (2011). Incorporating multiple criteria in HTA.

Dodgson, J. S., Spackman, M., Pearman, A., & Phillips, L. D. (2009). Multi-criteria analysis: a manual.

Domingo Galindo, L. (2016). *The Challenges of Logistics 4.0 for the Supply Chain Management and the Information Technology*. NTNU.

Foroozesh, N., Tavakkoli-Moghaddam, R., & Mousavi, S. M. (2017). Resilient supplier selection in a supply chain by a new interval-valued fuzzy group decision model based on possibilistic statistical concepts. *Journal of Industrial and Systems Engineering, 10*(2), 113-133.

Gan, J., Zhong, S., Liu, S., & Yang, D. (2019). Resilient Supplier Selection Based on Fuzzy BWM and GMo-RTOPSIS under Supply Chain Environment. *Discrete Dynamics in Nature and Society, 2019*, 14. doi: 10.1155/2019/2456260

Golden, B. L., Wasil, E. A., & Harker, P. T. (1989). The analytic hierarchy process. *Applications and Studies, Berlin, Heidelberg*.

Gregory, R., Failing, L., Harstone, M., Long, G., McDaniels, T., & Ohlson, D. (2012). *Structured decision making: a practical guide to environmental management choices*: John Wiley & Sons.

Guijun, W., & Xiaoping, L. (1998). The applications of interval-valued fuzzy numbers and interval-distribution numbers. *Fuzzy Sets and Systems, 98*(3), 331-335.

Guneri, A. F., Yucel, A., & Ayyildiz, G. (2009). An integrated fuzzy-lp approach for a supplier selection problem in supply chain management. *Expert Systems with Applications, 36*(5), 9223-9228.

Haibin, W., Smarandache, F., Zhang, Y., & Sunderraman, R. (2010). *Single valued neutrosophic sets*: Infinite Study.

Haldar, A., Ray, A., Banerjee, D., & Ghosh, S. (2012). A hybrid MCDM model for resilient supplier selection. *International Journal of Management Science and Engineering Management, 7*(4), 284-292.

Haldar, A., Ray, A., Banerjee, D., & Ghosh, S. (2014). Resilient supplier selection under a fuzzy environment. *International Journal of Management Science and Engineering Management, 9*(2), 147-156.

Hasan, M. M., Shohag, M. A. S., Azeem, A., Paul, S. K. J. I. J. o. L. S., & Management. (2015). Multiple criteria supplier selection: a fuzzy approach. *20*(4), 429-446.

Herrera, F., & Herrera-Viedma, E. (2000). Linguistic decision analysis: steps for solving decision problems under linguistic information. *Fuzzy Sets and Systems, 115*(1), 67-82. doi: 10.1016/S0165-0114(99)00024-X

Hofmann, E., & Rüsch, M. J. C. i. I. (2017). Industry 4.0 and the current status as well as future prospects on logistics. *89*, 23-34.

Hwang, C.-L., Lai, Y.-J., & Liu, T.-Y. (1993). A new approach for multiple objective decision making. *Computers & operations research, 20*(8), 889-899.

i-SCOOP. (2017a). Industry 4.0: the fourth industrial revolution – guide to Industrie 4.0. Retrieved 01/26, 2019, from https://www.i-scoop.eu/industry-4-0/

i-SCOOP. (2017b). Logistics 4.0 and smart supply chain management in Industry 4.0. Retrieved 02/26, 2019, from https://www.i-scoop.eu/industry-4-0/supply-chain-management-scm-logistics/#Logistics_40_the_crucial_aspect_of_autonomous_decisions_and_applications

Ivanov, D., Dolgui, A., Sokolov, B., Werner, F., & Ivanova, M. J. I. J. o. P. R. (2016). A dynamic model and an algorithm for short-term supply chain scheduling in the smart factory industry 4.0. *54*(2), 386-402.

Jiang, D., Faiz, T. I., & Hassan, M. (2018). *A Possibility Distribution Based Multi-Criteria Decision Algorithm for Resilient Supplier Selection Problems*: Infinite Study.

Juhász, J., & Bányai, T. (2018). *What Industry 4.0 Means for Just-In-Sequence Supply in Automotive Industry?* Paper presented at the Vehicle and Automotive Engineering.

Kamalahmadi, M., & Mellat-Parast, M. (2016). Developing a resilient supply chain through supplier flexibility and reliability assessment. *International Journal of Production Research, 54*(1), 302-321. doi: 10.1080/00207543.2015.1088971

Lakshmana Gomathi Nayagam, V., Muralikrishnan, S., & Sivaraman, G. (2011). Multi-criteria decision-making method based on interval-valued intuitionistic fuzzy sets. *Expert Systems with Applications, 38*(3), 1464-1467. doi: 10.1016/j.eswa.2010.07.055

Liao, C.-N., & Kao, H.-P. (2011). An integrated fuzzy TOPSIS and MCGP approach to supplier selection in supply chain management. *Expert Systems with Applications, 38*(9), 10803-10811. doi: 10.1016/j.eswa.2011.02.031

Lo, H.-W., & Liou, J. J. J. A. S. C. (2018). A novel multiple-criteria decision-making-based FMEA model for risk assessment. *73*, 684-696.

Mahapatra, G., Mahapatra, B., & Roy, P. (2016). A new concept for fuzzy variable based non-linear programming problem with application on system reliability via genetic algorithm approach. *Annals of Operations Research, 247*(2), 853-866. doi: 10.1007/s10479-015-1863-z

Mohammad, I. (2012). Group Decision Making Process for Supplier Selection with TOPSIS Method under Interval-Valued Intuitionistic Fuzzy Numbers. *Advances in Fuzzy Systems, 2012*(2012). doi: 10.1155/2012/407942

Mühlbacher, A. C., Kaczynski, A. J. A. h. e., & policy, h. (2016). Making good decisions in healthcare with multi-criteria decision analysis: the use, current research and future development of MCDA. *14*(1), 29-40.

Namdar, J., Li, X., Sawhney, R., & Pradhan, N. (2018). Supply chain resilience for single and multiple sourcing in the presence of disruption risks. *International Journal of Production Research, 56*(6), 2339-2360. doi: 10.1080/00207543.2017.1370149

Nutt, D. J., King, L. A., & Phillips, L. D. J. T. L. (2010). Drug harms in the UK: a multicriteria decision analysis. *376*(9752), 1558-1565.

Prasad, K., Prasad, M., Rao, S., & Patro, C. J. S. I. J. o. I. E. (2016). Supplier Selection through AHP-VIKOR Integrated Methodology. *3*(5), 1-6.

Rai, A., Patnayakuni, R., & Seth, N. J. M. q. (2006). Firm performance impacts of digitally enabled supply chain integration capabilities. 225-246.

Raiffa, H., & Keeney, R. (1975). Decision Analysis with Multiple Conflicting Objectives, Preferences and Value Tradeoffs.

Ren, Z., Xu, Z., & Wang, H. J. A. S. C. (2018). Multi-criteria group decision-making based on quasi-order for dual hesitant fuzzy sets and professional degrees of decision makers. *71*, 20-35.

Rozados, I. V., & Tjahjono, B. (2014). *Big data analytics in supply chain management: Trends and related research.* Paper presented at the 6th International Conference on Operations and Supply Chain Management, Bali.

Rüßmann, M., Lorenz, M., Gerbert, P., Waldner, M., Justus, J., Engel, P., & Harnisch, M. J. B. C. G. (2015). Industry 4.0: The future of productivity and growth in manufacturing industries. *9*(1), 54-89.

Saaty, T. L. J. N. Y. M. (1980). The analytic process: planning, priority setting, resources allocation.

Şahin, R., & Yiğider, M. (2014). A Multi-criteria neutrosophic group decision making metod based TOPSIS for supplier selection.

Shahroudi, K., & Tonekaboni, S. M. S. J. J. o. G. S. M. (2012). Application of TOPSIS method to supplier selection in Iran auto supply chain. *12*, 123-131.

Sharif Ullah, A. M. M. (2005). A fuzzy decision model for conceptual design. *Systems Engineering, 8*(4), 296-308. doi: 10.1002/sys.20038

Sharif Ullah, A. M. M., & Shamsuzzaman, M. (2013). Fuzzy Monte Carlo Simulation using point-cloud-based probability–possibility transformation. *SIMULATION, 89*(7), 860-875. doi: 10.1177/0037549713482174

Sheffi, Y., & Rice Jr, J. B. J. M. S. m. r. (2005). A supply chain view of the resilient enterprise: an organization's ability to recover from disruption quickly can be improved by building redundancy and flexibility into its supply chain. While investing in redundancy represents a pure cost increase, investing in flexibility yields many additional benefits for day-to-day operations. *47*(1), 41-49.

Sheffi, Y. J. M. P. B. (2005). The resilient enterprise: overcoming vulnerability for competitive advantage. *1*.

Singh, A. (2016). A goal programming approach for supplier evaluation and demand allocation among suppliers. *International Journal of Integrated Supply Management, 10*(1), 38-62.

Sodenkamp, M. A., Tavana, M., & Di Caprio, D. J. A. S. C. (2018). An aggregation method for solving group multi-criteria decision-making problems with single-valued neutrosophic sets. *71*, 715-727.

SUPPLYCHAINDIVE. (2018). Delivery issues jumped 50% as Florence hit. Retrieved 02/26, 2019, from https://www.supplychaindive.com/news/delivery-issues-jumped-50-as-florence-hit/533049/

Thokala, P., Devlin, N., Marsh, K., Baltussen, R., Boysen, M., Kalo, Z., . . . Ijzerman, M. (2016). Multiple Criteria Decision Analysis for Health Care Decision Making—An Introduction: Report 1 of the ISPOR MCDA Emerging Good Practices Task Force. *Value in Health, 19*(1), 1-13. doi: 10.1016/j.jval.2015.12.003

Torabi, S. A., Baghersad, M., & Mansouri, S. A. (2015). Resilient supplier selection and order allocation under operational and disruption risks. *Transportation Research Part E: Logistics and Transportation Review, 79*, 22-48. doi: https://doi.org/10.1016/j.tre.2015.03.005

Ullah, A. M. M. S. (2005). Handling design perceptions: an axiomatic design perspective. *Research in Engineering Design, 16*(3), 109-117. doi: 10.1007/s00163-005-0002-2

Ullah, A. M. M. S., & Noor-E-Alam, M. (2018). Big data driven graphical information based fuzzy multi criteria decision making. *Applied Soft Computing, 63*, 23-38. doi: 10.1016/j.asoc.2017.11.026

Valipour Parkouhi, S., Safaei Ghadikolaei, A., & Fallah Lajimi, H. (2019). Resilient supplier selection and segmentation in grey environment. *Journal of Cleaner Production, 207*, 1123-1137. doi: https://doi.org/10.1016/j.jclepro.2018.10.007

Wahlster, P., Goetghebeur, M., Kriza, C., Niederländer, C., & Kolominsky-Rabas, P. J. B. h. s. r. (2015). Balancing costs and benefits at different stages of medical innovation: a systematic review of Multi-criteria decision analysis (MCDA). *15*(1), 262.

Waller, M. A., & Fawcett, S. E. J. J. o. B. L. (2013). Data science, predictive analytics, and big data: a revolution that will transform supply chain design and management. *34*(2), 77-84.

Wang, G., Gunasekaran, A., Ngai, E. W., & Papadopoulos, T. J. I. J. o. P. E. (2016). Big data analytics in logistics and supply chain management: Certain investigations for research and applications. *176*, 98-110.

Wang, H., Smarandache, F., Sunderraman, R., & Zhang, Y.-Q. (2005). *interval neutrosophic sets and logic: theory and applications in computing: Theory and applications in computing* (Vol. 5): Infinite Study.

Wen, J., Miaoyan, Z., & Chunhe, X. (2017). A Reliability-Based Method to Sensor Data Fusion. *Sensors, 17*(7), 1575. doi: 10.3390/s17071575

Witkowski, K. J. P. E. (2017). Internet of things, big data, industry 4.0–innovative solutions in logistics and supply chains management. *182*, 763-769.

Ye, L., & Abe, M. (2012). The impacts of natural disasters on global supply chains: ARTNeT working paper series.

Yoon, K. (1987). A reconciliation among discrete compromise solutions. *Journal of the Operational Research Society, 38*(3), 277-286.

Zadeh, L. A. (1965). Fuzzy sets. *Information and Control, 8*(3), 338-353. doi: https://doi.org/10.1016/S0019-9958(65)90241-X

Zadeh, L. A. (1997). Toward a theory of fuzzy information granulation and its centrality in human reasoning and fuzzy logic. *Fuzzy Sets and Systems, 90*(2), 111-127. doi: https://doi.org/10.1016/S0165-0114(97)00077-8

Zhong, R. Y., Xu, X., Klotz, E., & Newman, S. T. J. E. (2017). Intelligent manufacturing in the context of industry 4.0: a review. *3*(5), 616-630.

# Appendix A

## Time-series Data

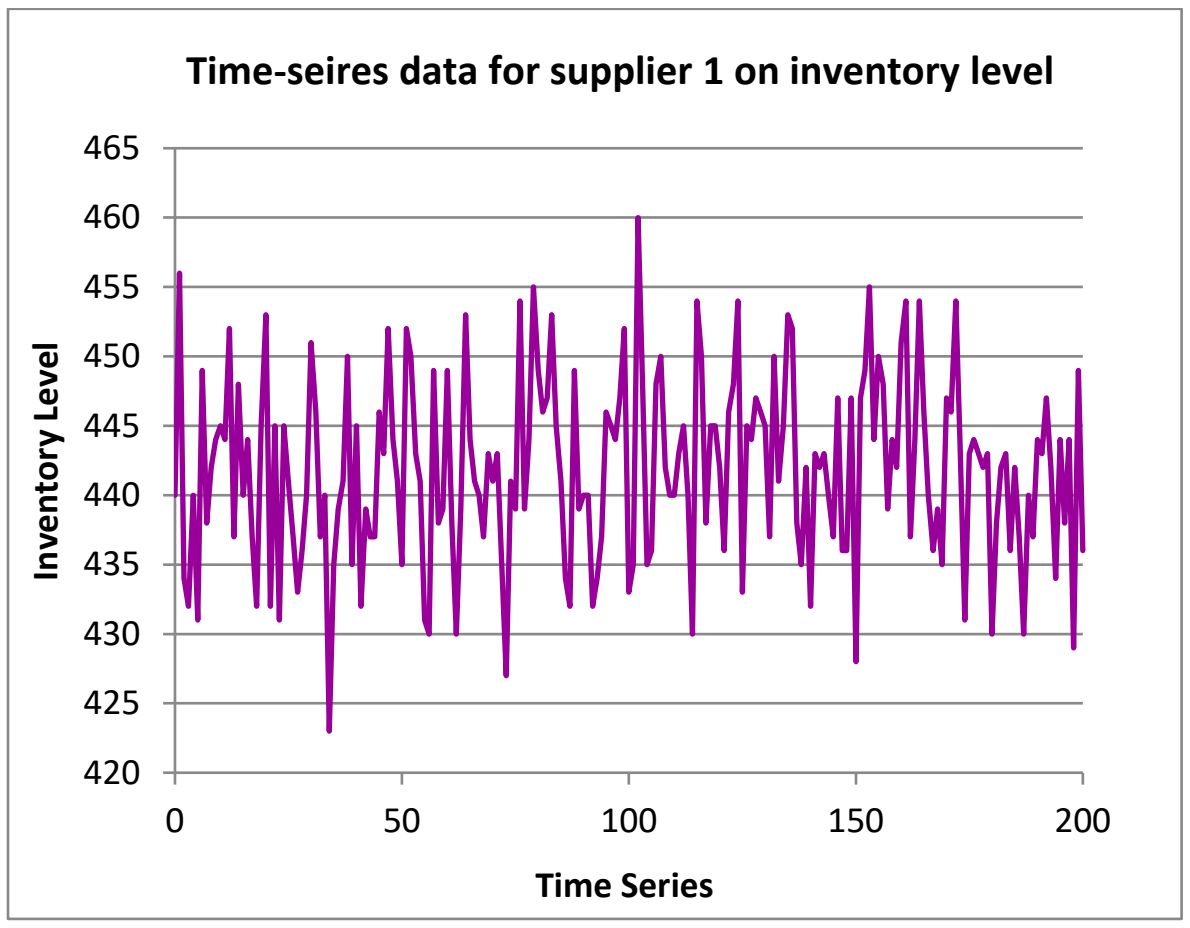

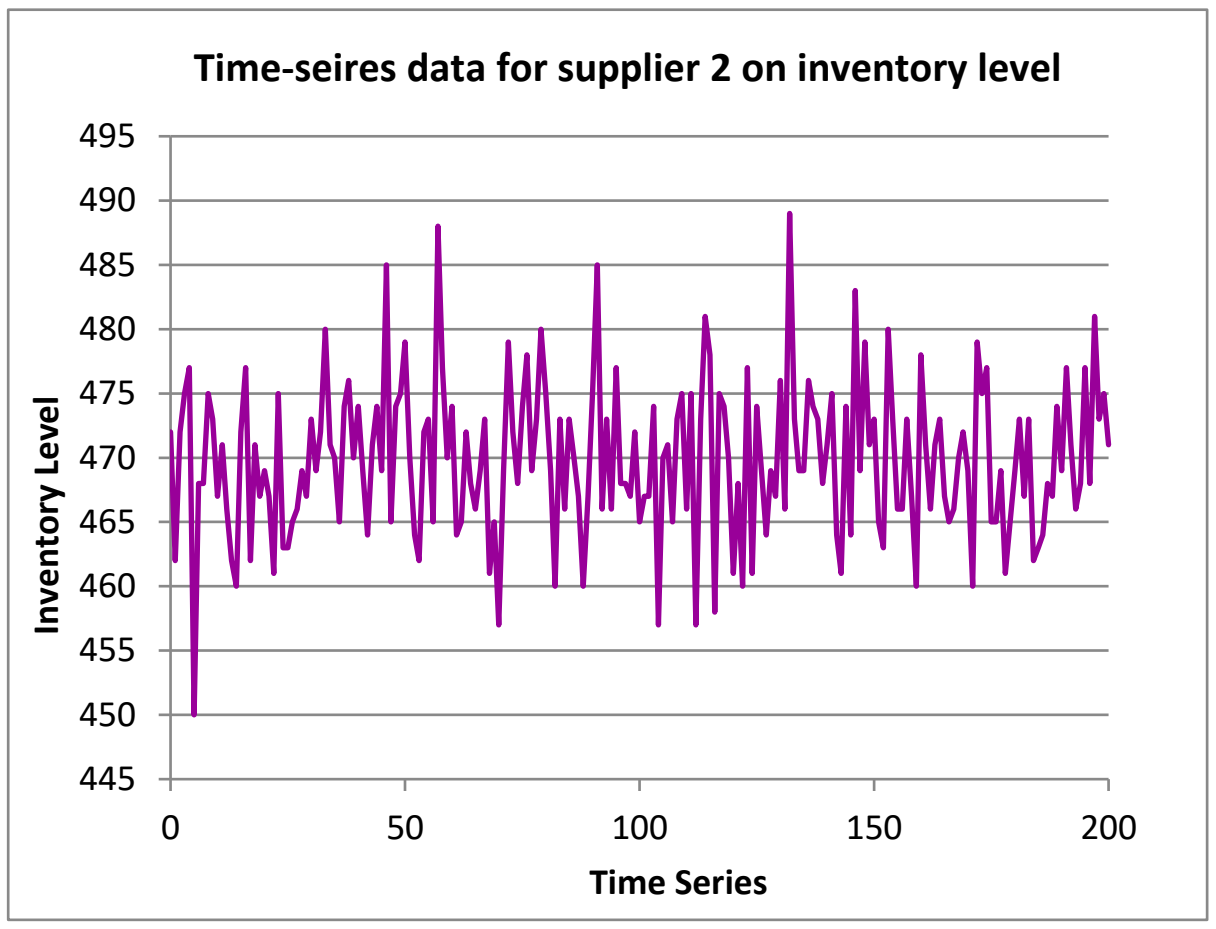

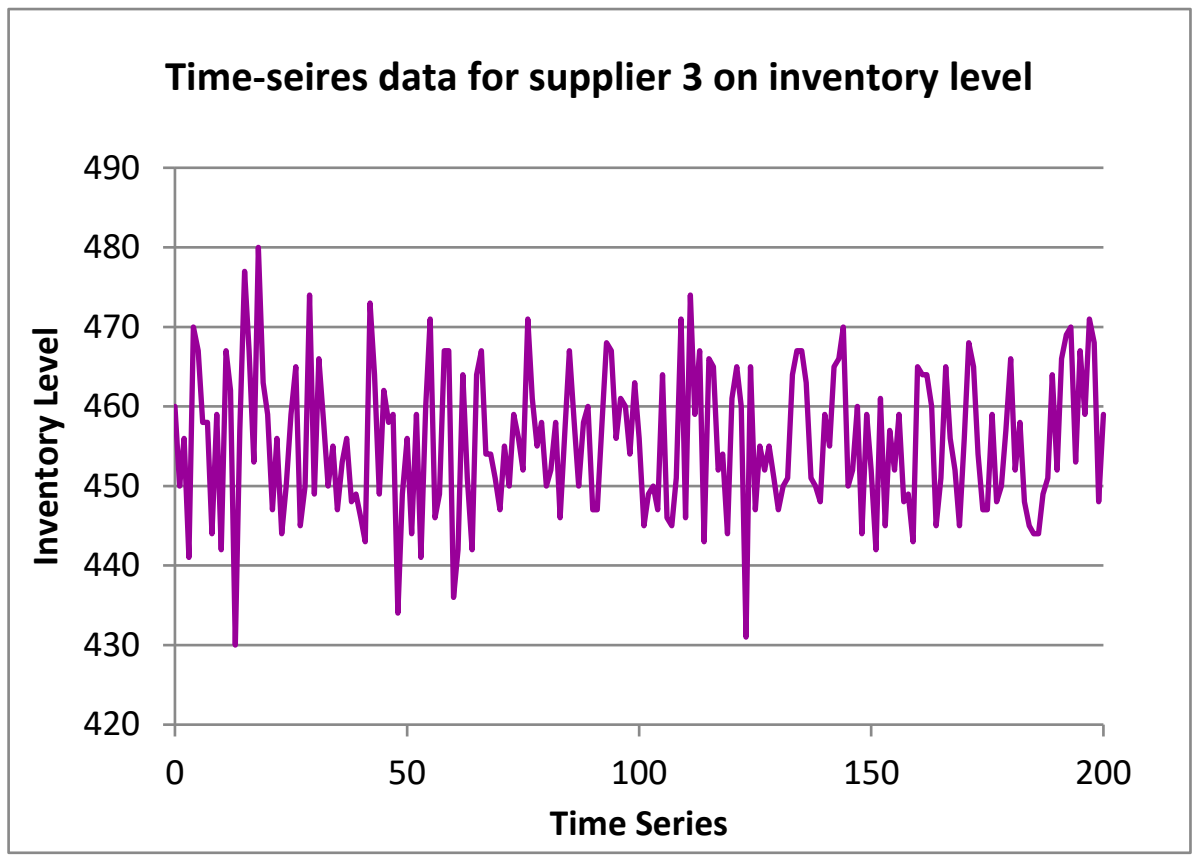

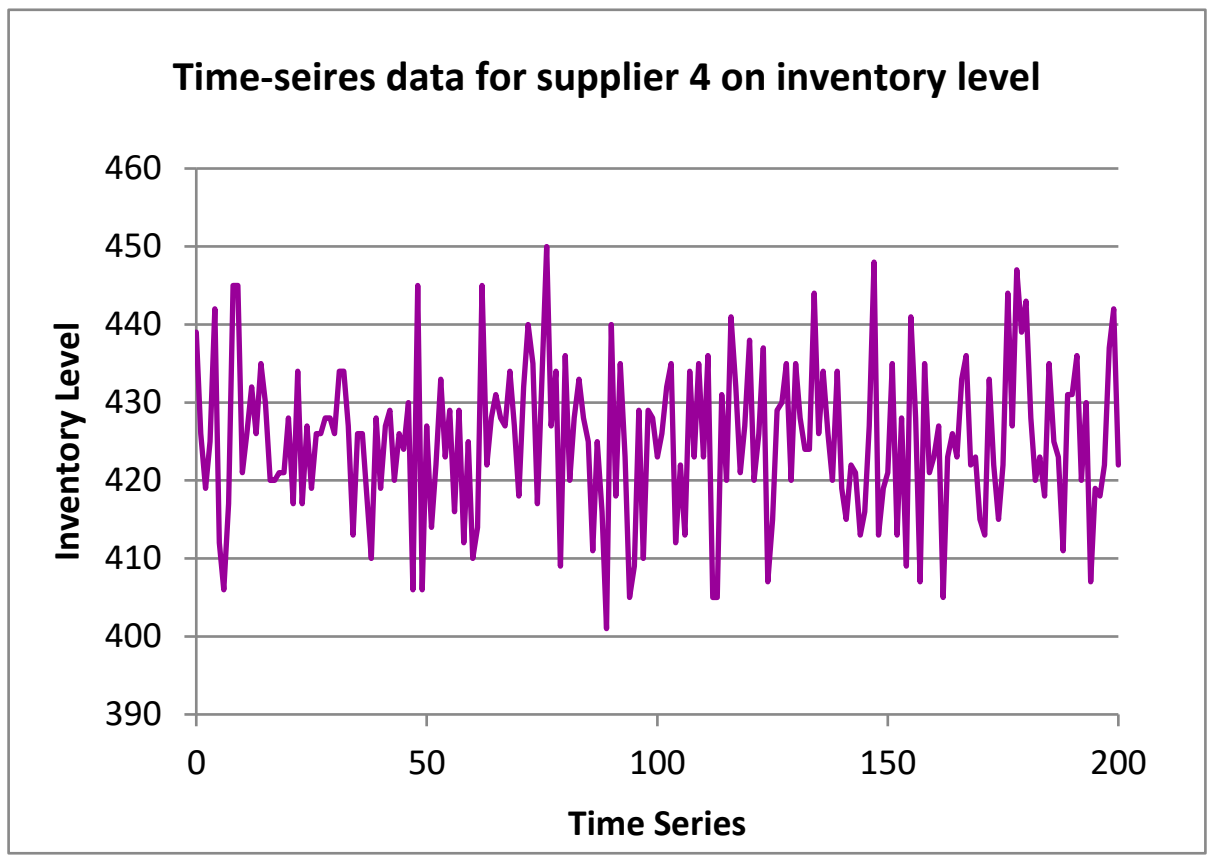

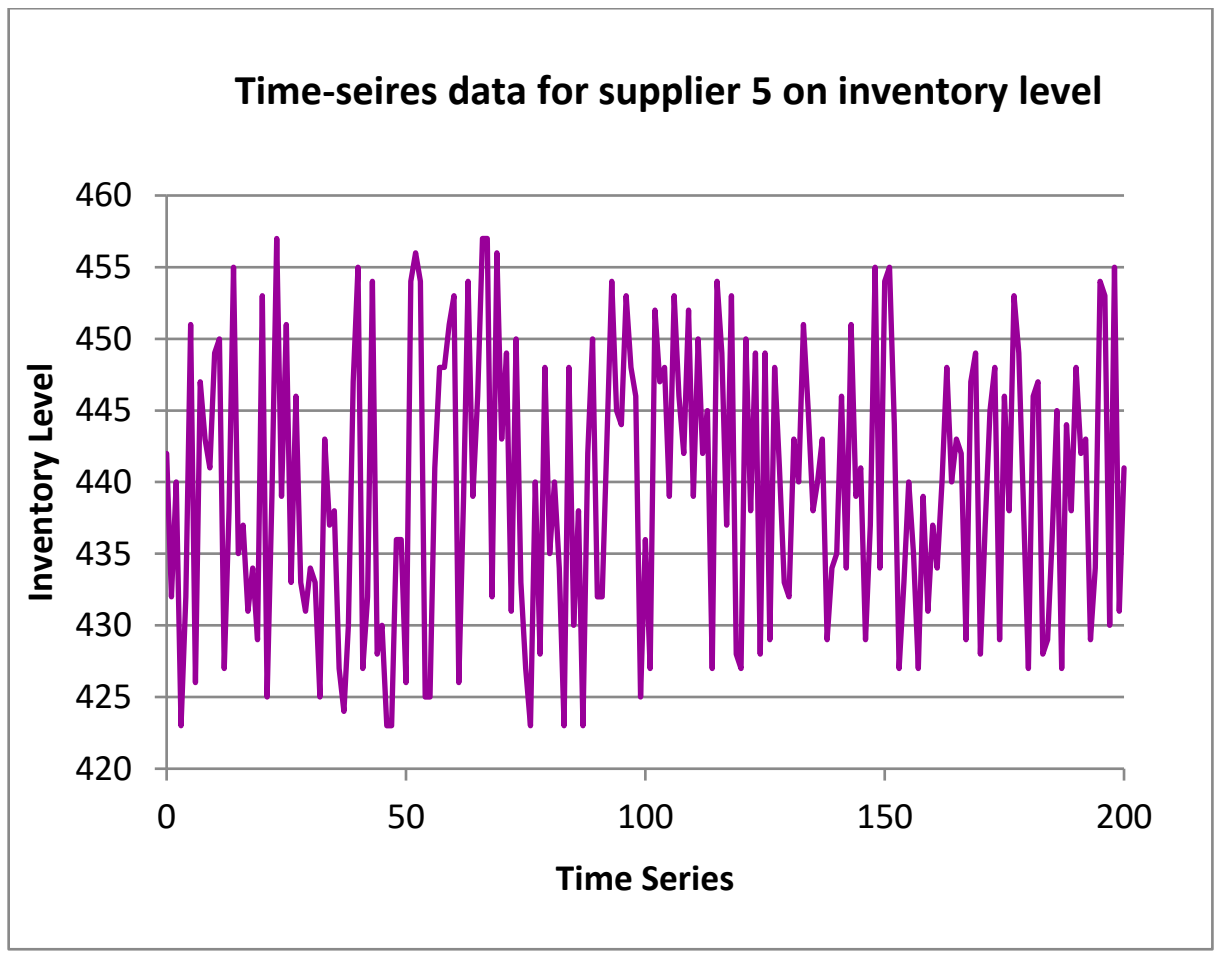

Figure 15. Time-series data for suppliers on pre-positioned inventory level ($C_1$)

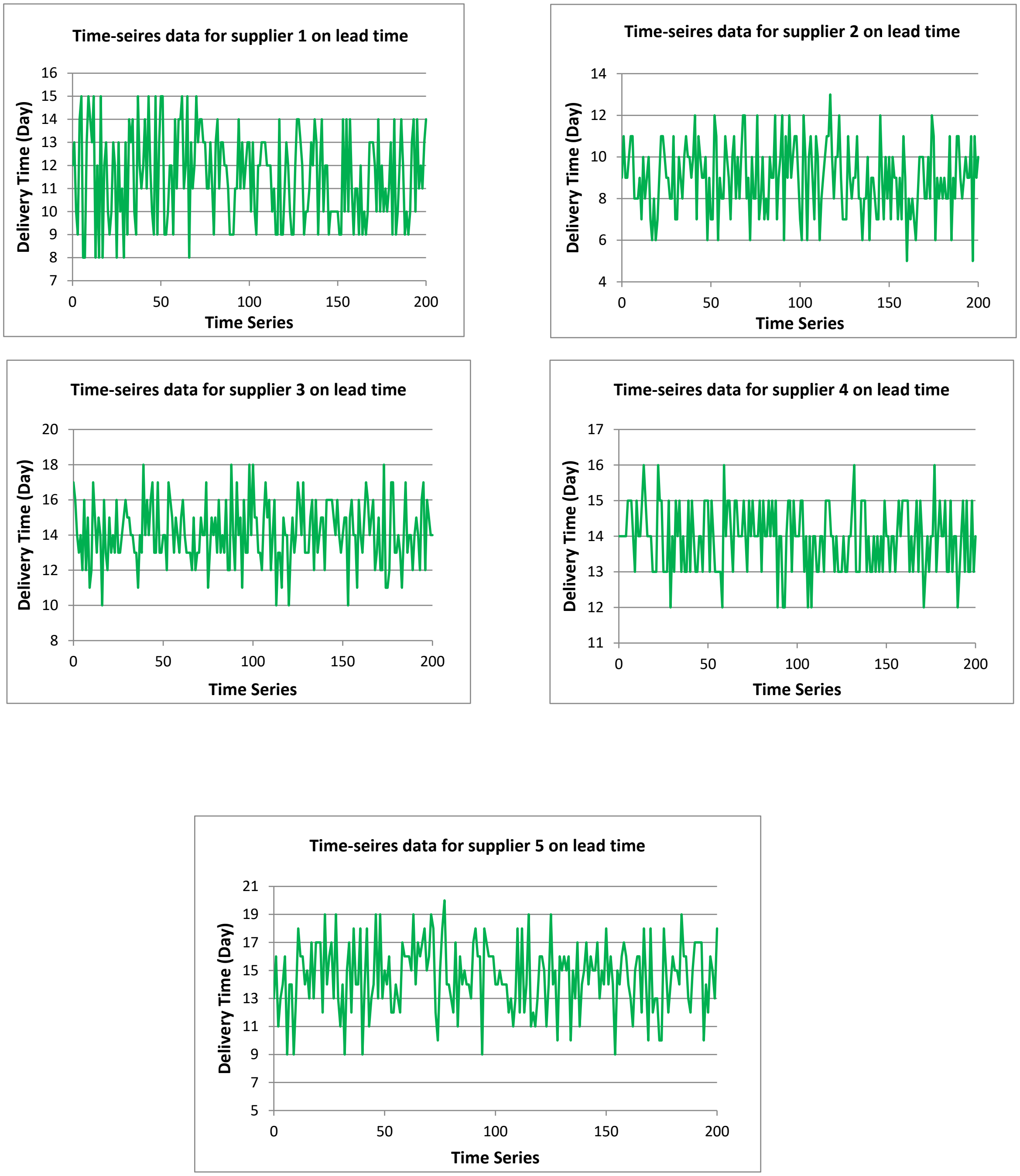

Figure 16. Time-series data for suppliers on lead time variability ($C_2$)

**Historically collected data presented in graphical format**

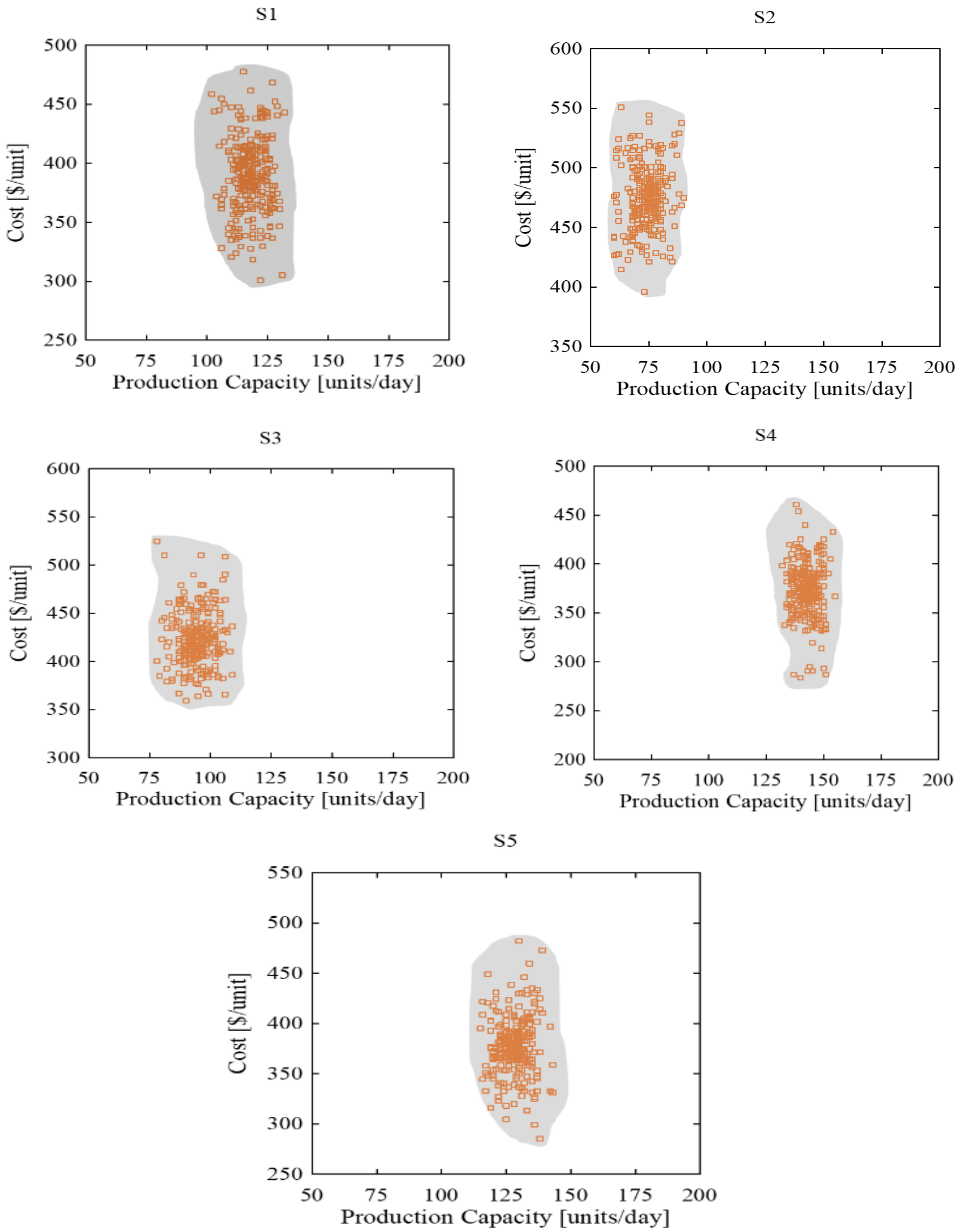

Figure 17. Graphical information for suppliers on non-time-series attributes

SCRI values created for different $\alpha$ values and normalized $\tilde{\rho}_i$ according to equation 6.1

Table 12
SCRI based on $\alpha$

| $Supplier/\alpha$ | 0.1 | 0.2 | 0.3 | 0.4 | 0.5 | 0.6 | 0.7 | 0.8 | 0.9 |
|---|---|---|---|---|---|---|---|---|---|
| $S_1$ | 0.207 | 0.207 | 0.206 | 0.206 | 0.206 | 0.205 | 0.205 | 0.205 | 0.204 |
| $S_2$ | 0.179 | 0.182 | 0.185 | 0.189 | 0.192 | 0.195 | 0.198 | 0.201 | 0.204 |
| $S_3$ | 0.194 | 0.196 | 0.198 | 0.200 | 0.203 | 0.205 | 0.207 | 0.209 | 0.212 |
| $S_4$ | 0.212 | 0.210 | 0.208 | 0.206 | 0.204 | 0.202 | 0.200 | 0.198 | 0.195 |
| $S_5$ | 0.208 | 0.205 | 0.202 | 0.199 | 0.196 | 0.193 | 0.190 | 0.187 | 0.184 |

**Linguistic Data:**

Table 13.
Linguistic data (Performance measurement and attribute weight)

| $C_5$ (Digitalization) | | | | | |
|---|---|---|---|---|---|
| Supplier | $DM_1$ | $DM_2$ | $DM_3$ | $DM_4$ | $DM_5$ |
| $S_1$ | VG | VG | VVG | VG | G |
| $S_2$ | VG | VVG | EG | VG | VG |
| $S_3$ | EG | VG | VVG | VG | EG |
| $S_4$ | VG | VG | G | MG | VVG |
| $S_5$ | M | MG | MG | MG | MB |

| $C_6$ (Traceability) | | | | | |
|---|---|---|---|---|---|
| Supplier | $DM_1$ | $DM_2$ | $DM_3$ | $DM_4$ | $DM_5$ |
| $S_1$ | VG | VG | VVG | VG | VVG |
| $S_2$ | VG | VG | G | MG | MG |
| $S_3$ | EG | VVG | VVG | EG | VG |
| $S_4$ | MG | M | M | MG | G |
| $S_5$ | M | M | MG | MG | G |

| $C_7$ (Supply Chain Density) | | | | | |
|---|---|---|---|---|---|
| Supplier | $DM_1$ | $DM_2$ | $DM_3$ | $DM_4$ | $DM_5$ |
| $S_1$ | VG | VG | G | MG | VVG |
| $S_2$ | M | MG | MG | MG | MG |
| $S_3$ | G | VG | VVG | VG | G |
| $S_4$ | VG | VVG | VVG | EG | EG |
| $S_5$ | VG | VVG | VG | VG | VVG |

| $C_8$ (Supply Chain Complexity) | | | | | |
|---|---|---|---|---|---|
| Supplier | $DM_1$ | $DM_2$ | $DM_3$ | $DM_4$ | $DM_5$ |
| $S_1$ | G | G | VG | VG | G |
| $S_2$ | VVG | VG | VG | VVG | VG |
| $S_3$ | M | G | MG | MG | G |
| $S_4$ | VG | VVG | VVG | VVG | VG |
| $S_5$ | M | MG | MG | MG | M |

| $C_9$ (Re-engineering) | | | | | |
|---|---|---|---|---|---|
| Supplier | $DM_1$ | $DM_2$ | $DM_3$ | $DM_4$ | $DM_5$ |
| $S_1$ | M | MG | M | M | MG |
| $S_2$ | M | MB | MB | MB | B |
| $S_3$ | G | VG | G | G | G |
| $S_4$ | M | MG | M | M | M |
| $S_5$ | M | M | MG | MG | MB |

| $C_{10}$(Supplier's Resource Flexibility) | | | | | |
|---|---|---|---|---|---|
| Supplier | $DM_1$ | $DM_2$ | $DM_3$ | $DM_4$ | $DM_5$ |
| $S_1$ | VG | VVG | VVG | VG | VG |
| $S_2$ | M | MG | G | M | MG |
| $S_3$ | VG | VVG | VVG | VG | VG |
| $S_4$ | VG | VVG | VVG | VG | EG |
| $S_5$ | G | G | M | MG | G |

| $C_{11}$ (Automation Disruption) | | | | | |
|---|---|---|---|---|---|
| Supplier | $DM_1$ | $DM_2$ | $DM_3$ | $DM_4$ | $DM_5$ |
| $S_1$ | G | G | VVG | G | VG |
| $S_2$ | VG | G | VVG | VG | VG |
| $S_3$ | M | B | M | M | B |
| $S_4$ | MG | MG | M | G | G |
| $S_5$ | EG | VVG | VVG | VG | VVG |

| $C_{12}$(Information Management) | | | | | |
|---|---|---|---|---|---|
| Supplier | $DM_1$ | $DM_2$ | $DM_3$ | $DM_4$ | $DM_5$ |
| $S_1$ | MG | MG | M | G | G |
| $S_2$ | MG | G | G | MG | MG |
| $S_3$ | VG | G | VG | VG | G |
| $S_4$ | VG | VG | VVG | VVG | VG |
| $S_5$ | MG | MG | G | M | M |

| $C_{13}$ (Cyber security Risk Management) | | | | | |
|---|---|---|---|---|---|
| Supplier | $DM_1$ | $DM_2$ | $DM_3$ | $DM_4$ | $DM_5$ |
| $S_1$ | MG | MG | G | M | MG |
| $S_2$ | G | VG | VG | G | G |
| $S_3$ | VG | VG | VVG | VVG | VG |
| $S_4$ | MG | MG | MG | M | MG |
| $S_5$ | VG | G | G | MG | MG |

| $C_{14}$(Supplier reliability) | | | | | |
|---|---|---|---|---|---|
| Supplier | $DM_1$ | $DM_2$ | $DM_3$ | $DM_4$ | $DM_5$ |
| $S_1$ | EG | VG | VVG | VVG | EG |
| $S_2$ | MG | MG | G | MG | MG |
| $S_3$ | VG | MG | G | G | VG |
| $S_4$ | M | M | G | G | MG |
| $S_5$ | G | VG | G | G | VG |

| $C_{15}$ (Supply chain visibility) | | | | | |
|---|---|---|---|---|---|
| Supplier | $DM_1$ | $DM_2$ | $DM_3$ | $DM_4$ | $DM_5$ |
| $S_1$ | VG | VG | G | G | G |
| $S_2$ | VG | G | VVG | VG | VG |
| $S_3$ | G | G | G | VG | G |
| $S_4$ | MG | M | M | MG | MG |
| $S_5$ | MG | M | M | MB | M |

| $C_{16}$(Level of collaboration) | | | | | |
|---|---|---|---|---|---|
| Supplier | $DM_1$ | $DM_2$ | $DM_3$ | $DM_4$ | $DM_5$ |
| $S_1$ | G | G | VG | G | G |
| $S_2$ | VG | VVG | EG | VVG | EG |
| $S_3$ | G | G | VG | G | G |
| $S_4$ | VG | VG | G | G | MG |
| $S_5$ | VVG | VVG | VG | VG | VVG |

| $C_{17}$ (Restorative capacity) | | | | | |
|---|---|---|---|---|---|
| Supplier | $DM_1$ | $DM_2$ | $DM_3$ | $DM_4$ | $DM_5$ |
| $S_1$ | M | G | G | MG | MG |
| $S_2$ | G | VG | VG | G | VG |
| $S_3$ | EG | VG | VVG | VG | VVG |
| $S_4$ | M | M | MG | MG | M |
| $S_5$ | VVG | G | VG | G | G |

| $C_{18}$(Rerouting) | | | | | |
|---|---|---|---|---|---|
| Supplier | $DM_1$ | $DM_2$ | $DM_3$ | $DM_4$ | $DM_5$ |
| $S_1$ | G | VG | G | G | VG |
| $S_2$ | VG | VVG | VG | EG | VVG |
| $S_3$ | G | MG | G | M | MG |
| $S_4$ | VG | G | VG | VG | G |
| $S_5$ | G | G | MG | G | MG |

| $C_{19}$ (Agility) | | | | | |
|---|---|---|---|---|---|
| Supplier | $DM_1$ | $DM_2$ | $DM_3$ | $DM_4$ | $DM_5$ |
| $S_1$ | VG | G | G | MG | G |
| $S_2$ | M | MG | G | G | MG |
| $S_3$ | G | MG | G | MG | MG |
| $S_4$ | VG | G | VG | G | G |
| $S_5$ | G | VG | VG | G | VG |

| Weight of the attribute | | | | | |
|---|---|---|---|---|---|
| Attribute | $DM_1$ | $DM_2$ | $DM_3$ | $DM_4$ | $DM_5$ |
| $C_1$ | I | VI | EI | I | VI |
| $C_2$ | M | I | VI | I | M |
| $C_3$ | M | M | I | M | I |
| $C_4$ | I | I | VI | I | I |
| $C_5$ | I | I | VI | VI | I |
| $C_6$ | EI | I | EI | EI | I |
| $C_7$ | M | M | I | I | M |
| $C_8$ | I | VI | I | I | VI |
| $C_9$ | M | UI | M | M | UI |
| $C_{10}$ | I | VI | VI | VI | I |
| $C_{11}$ | I | I | VI | VI | I |
| $C_{12}$ | I | MI | MI | I | MI |
| $C_{13}$ | M | MI | I | MI | I |
| $C_{14}$ | VI | I | I | VI | VI |
| $C_{15}$ | MI | I | M | MI | M |
| $C_{16}$ | I | VI | VI | MI | I |
| $C_{17}$ | I | M | M | VI | I |
| $C_{18}$ | I | I | MI | M | MI |
| $C_{19}$ | M | MI | I | MI | MI |

Table 14.
Linguistic terms (used in performance measurement) and coresponding TFNs

| Performance Measurement | |
|---|---|
| Linguistic Terms | TFN (a, b, c) |
| VB | (0,1,2) |
| B | (1,2,3) |
| … | … |
| EG | (8,9,10) |

# Appendix B

## Attributes concerning time series data

Based on the approach described in sub-step 1(a) and the original time-series data in Figure 15 and Figure 16, the possibility distribution in the form of triangular fuzzy number are generated as follows:

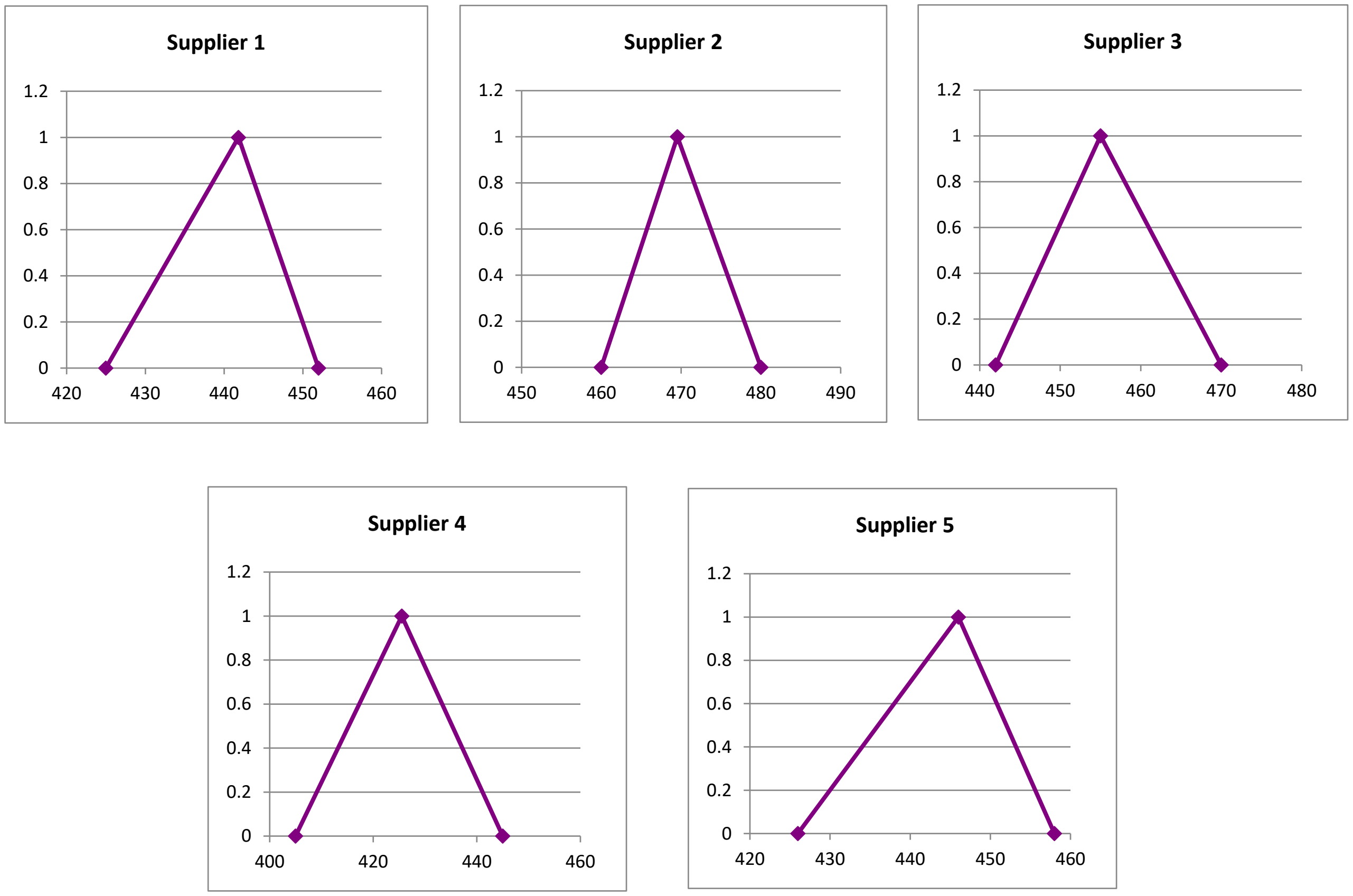

Figure 18. Graphical Triangular Fuzzy Number for suppliers on pre-positioned inventory level ($C_1$)

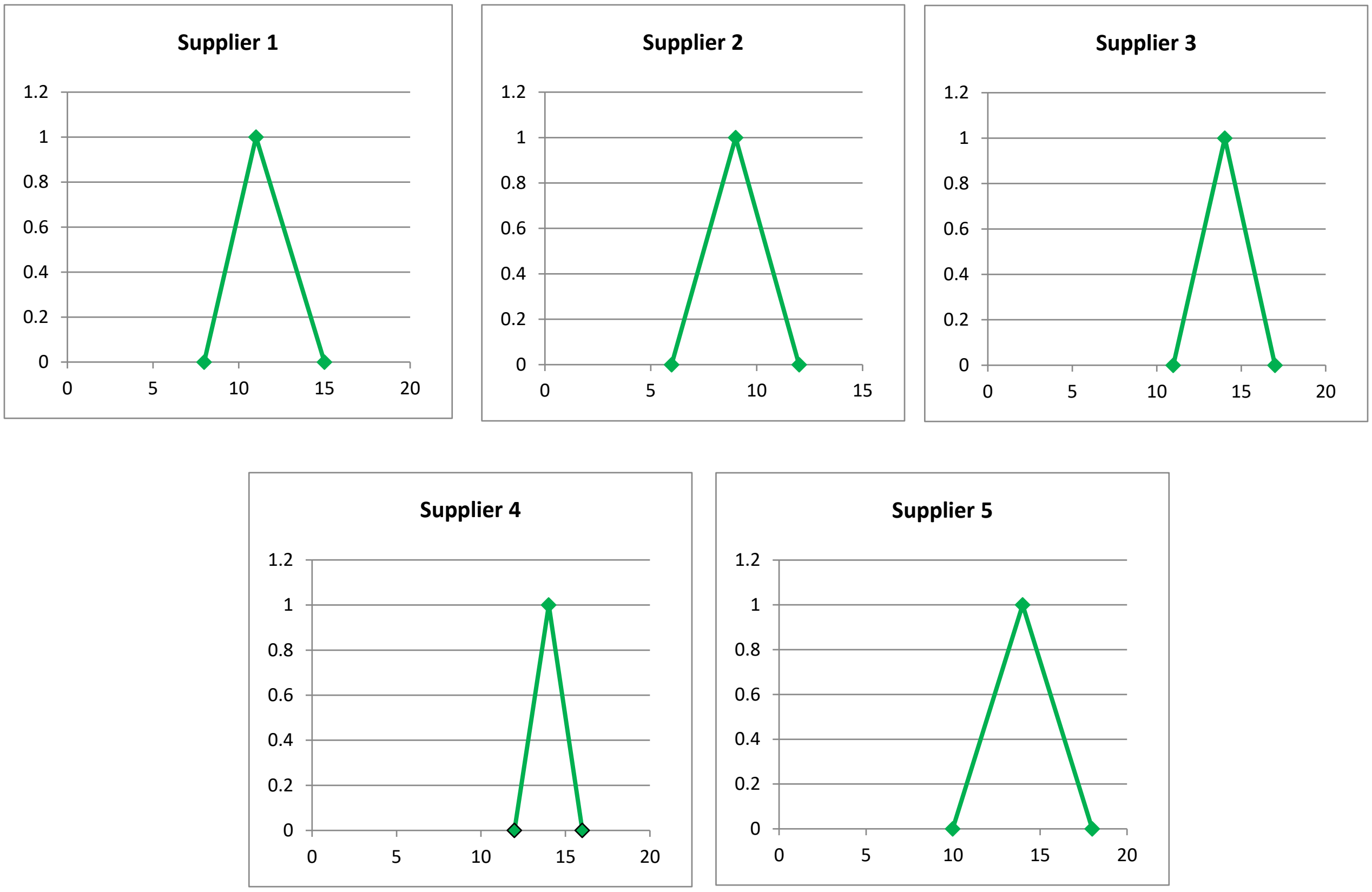

Figure 19. Graphical Triangular Fuzzy Number for suppliers on lead time variability ($C_2$)

With the obtained graphical TFN, the numerical TFN could be constructed from Figure 18 and Figure 19 as:

Table 15.
Numerical Triangular Fuzzy Number for suppliers on $C_1$ and $C_2$

| Supplier | $C_1$ (Pre-positioned Inventory Level) | | | $C_2$ (Lead Time Variability) | | |
|---|---|---|---|---|---|---|
| | a | b | c | a | b | c |
| $S_1$ | 423.98 | 441.04 | 454 | 7.98 | 11.01 | 15 |
| $S_2$ | 459.98 | 469.54 | 480 | 5.98 | 9 | 12 |
| $S_3$ | 441.98 | 455.01 | 470 | 10.98 | 14.03 | 17 |
| $S_4$ | 404.98 | 425.51 | 445 | 11.98 | 14.01 | 16 |
| $S_5$ | 424.98 | 446.05 | 459 | 9.98 | 14 | 18 |

**Non-time-series Criteria:**

Based on the computation process summarized in sub-step 1(b) and non-time series data in Figure 17, the original data are extracted in the form of range value as follows:

Table 16.
Extracted range values for suppliers on non-time-series attributes

| $S_i$ | $S_1$ | | | | $S_2$ | | | | $S_3$ | | | | $S_4$ | | | | $S_5$ | | | |
|---|---|---|---|---|---|---|---|---|---|---|---|---|---|---|---|---|---|---|---|---|
| $C_i$ | $C_1$ | | $C_2$ | | $C_1$ | | $C_2$ | | $C_1$ | | $C_2$ | | $C_1$ | | $C_2$ | | $C_1$ | | $C_2$ | |
| $R_i$ | min | max | min | max | min | max | min | max | min | max | min | max | min | max | min | max | min | max | min | max |
| $R_1$ | 103 | 134 | 337 | 472 | 60 | 88 | 405 | 554 | 76 | 103 | 371 | 529 | 131 | 147 | 272 | 467 | 115 | 143 | 348 | 463 |
| $R_2$ | 96 | 136 | 310 | 478 | 60 | 90 | 400 | 555 | 76 | 110 | 361 | 531 | 128 | 152 | 345 | 455 | 112 | 144 | 314 | 474 |
| $R_3$ | 96 | 134 | 296 | 483 | 58 | 90 | 394 | 556 | 78 | 113 | 355 | 528 | 125 | 157 | 394 | 446 | 112 | 145 | 294 | 483 |
| $R_4$ | 96 | 135 | 295 | 483 | 59 | 89 | 391 | 556 | 79 | 113 | 351 | 528 | 128 | 158 | 270 | 458 | 111 | 144 | 283 | 489 |
| $R_5$ | 98 | 136 | 297 | 481 | 59 | 91 | 391 | 554 | 78 | 115 | 351 | 524 | 129 | 156 | 273 | 450 | 112 | 145 | 279 | 484 |
| $R_6$ | 100 | 136 | 300 | 478 | 60 | 91 | 391 | 554 | 77 | 115 | 354 | 522 | 129 | 157 | 283 | 442 | 111 | 146 | 282 | 478 |
| $R_7$ | 101 | 135 | 385 | 459 | 58 | 89 | 408 | 551 | 75 | 115 | 354 | 518 | 130 | 158 | 317 | 436 | 111 | 147 | 306 | 465 |
| $R_8$ | 103 | 135 | 300 | 473 | 58 | 88 | 420 | 547 | 75 | 113 | 357 | 511 | 131 | 157 | 326 | 462 | 111 | 148 | 311 | 382 |
| $R_9$ | 105 | 135 | 325 | 475 | 58 | 88 | 426 | 544 | 75 | 114 | 367 | 505 | 131 | 158 | 270 | 463 | 113 | 148 | 301 | 480 |
| $R_{10}$ | 110 | 136 | 301 | 482 | 60 | 87 | 430 | 536 | 75 | 113 | 363 | 529 | 134 | 156 | 270 | 454 | 114 | 149 | 287 | 486 |

With the extracted range values and the fuzzified frame of discernment presented in Figure 9, the TFN decision matrix is constructed as in Table 17:

Table 17.
Integrated Triangular Fuzzy Number for suppliers on non-time-series attributes

| Supplier | $C_3$ | | | $C_4$ | | |
|---|---|---|---|---|---|---|
| | a | b | c | a | b | c |
| $S_1$ | 102.98 | 117.23 | 131.49 | 320.06 | 394.96 | 488.16 |
| $S_2$ | 66.14 | 75.55 | 92.23 | 358.10 | 470.87 | 526.03 |
| $S_3$ | 81.25 | 95.04 | 109.53 | 338.87 | 432.49 | 506.88 |
| $S_4$ | 124.33 | 141.10 | 150.33 | 311.70 | 379.58 | 481.14 |
| $S_5$ | 113.44 | 128.67 | 140.98 | 316.41 | 385.76 | 482.62 |

According to sub-step 2(b) and Table 13, the attribute weight decision matrix is constructed as:

Table 18.
Attribute weight decision matrix

| Attribute | Weight | | |
|---|---|---|---|
| | a | b | c |
| $C_1$ | 0.5 | 0.7 | 1 |
| $C_2$ | 0.3 | 0.54 | 0.8 |
| $C_3$ | 0.3 | 0.48 | 0.7 |
| $C_4$ | 0.5 | 0.62 | 0.8 |
| $C_5$ | 0.5 | 0.64 | 0.8 |
| $C_6$ | 0.5 | 0.78 | 1 |
| $C_7$ | 0.3 | 0.48 | 0.7 |
| $C_8$ | 0.5 | 0.64 | 0.8 |
| $C_9$ | 0.1 | 0.32 | 0.5 |
| $C_{10}$ | 0.5 | 0.66 | 0.8 |
| $C_{11}$ | 0.5 | 0.64 | 0.8 |
| $C_{12}$ | 0.4 | 0.54 | 0.7 |
| $C_{13}$ | 0.3 | 0.52 | 0.7 |
| $C_{14}$ | 0.5 | 0.66 | 0.8 |
| $C_{15}$ | 0.3 | 0.48 | 0.7 |
| $C_{16}$ | 0.4 | 0.62 | 0.8 |
| $C_{17}$ | 0.3 | 0.54 | 0.8 |
| $C_{18}$ | 0.3 | 0.52 | 0.7 |
| $C_{19}$ | 0.3 | 0.5 | 0.7 |

The PIS and NIS matrix are generated based on function (5.6) and (5.7) in step 3:

Table 19.
PIS and NIS for attributes

| Attribute | PIS | | | NIS | | |
|---|---|---|---|---|---|---|
| TFN | a | b | c | a | b | c |
| $C_1$ | 1.00 | 1.00 | 1.00 | 0.42 | 0.42 | 0.42 |
| $C_2$ | 0.80 | 0.80 | 0.80 | 0.10 | 0.10 | 0.10 |
| $C_3$ | 0.70 | 0.70 | 0.70 | 0.06 | 0.06 | 0.06 |
| $C_4$ | 0.80 | 0.80 | 0.80 | 0.15 | 0.15 | 0.15 |
| $C_5$ | 0.80 | 0.80 | 0.80 | 0.22 | 0.22 | 0.22 |
| $C_6$ | 1.00 | 1.00 | 1.00 | 0.30 | 0.30 | 0.30 |
| $C_7$ | 0.70 | 0.70 | 0.70 | 0.09 | 0.09 | 0.09 |
| $C_8$ | 0.80 | 0.80 | 0.80 | 0.17 | 0.17 | 0.17 |
| $C_9$ | 0.50 | 0.50 | 0.50 | 0.01 | 0.01 | 0.01 |
| $C_{10}$ | 0.80 | 0.80 | 0.80 | 0.15 | 0.15 | 0.15 |
| $C_{11}$ | 0.80 | 0.80 | 0.80 | 0.05 | 0.05 | 0.05 |
| $C_{12}$ | 0.70 | 0.70 | 0.70 | 0.13 | 0.13 | 0.13 |
| $C_{13}$ | 0.70 | 0.70 | 0.70 | 0.10 | 0.10 | 0.10 |
| $C_{14}$ | 0.80 | 0.80 | 0.80 | 0.15 | 0.15 | 0.15 |
| $C_{15}$ | 0.70 | 0.70 | 0.70 | 0.07 | 0.07 | 0.07 |
| $C_{16}$ | 0.80 | 0.80 | 0.80 | 0.16 | 0.16 | 0.16 |
| $C_{17}$ | 0.80 | 0.80 | 0.80 | 0.09 | 0.09 | 0.09 |
| $C_{18}$ | 0.70 | 0.70 | 0.70 | 0.09 | 0.09 | 0.09 |
| $C_{19}$ | 0.7 | 0.7 | 0.7 | 0.11 | 0.11 | 0.11 |